\documentclass[journal,onecolumn,draftclsnofoot,]{IEEEtran}
\usepackage{multirow}
\usepackage{cite}
\usepackage[implicit=false]{hyperref}
\ifCLASSINFOpdf
  \usepackage[pdftex]{graphicx}
\else
  \usepackage[dvips]{graphicx}
  \DeclareGraphicsExtensions{.eps}
\fi
\usepackage{amsmath}
\usepackage{array}
\usepackage{amsthm}
\ifCLASSOPTIONcompsoc
  \usepackage[caption=false,font=normalsize,labelfont=sf,textfont=sf]{subfig}
\else
  \usepackage[caption=false,font=footnotesize]{subfig}
\fi

\begin{document}
\title{Activation Learning by Local Competitions}


\author{Hongchao~Zhou
%        John~Doe,~\IEEEmembership{Fellow,~OSA,}
%        and~Jane~Doe,~\IEEEmembership{Life~Fellow,~IEEE}% <-this % stops a space
\thanks{H. Zhou is with the School of Information Science and Engineering,
Shandong University, Qingdao, Shandong, 266237 China. Email:
hongchao@sdu.edu.cn.}}% <-this % stops a space
\maketitle

\begin{abstract}
Despite its great success, backpropagation has certain limitations that necessitate the investigation of new learning methods. In this study, we present a biologically plausible local learning rule that improves upon Hebb's well-known proposal and discovers unsupervised features by local competitions among neurons. This simple learning rule enables the creation of a forward learning paradigm called activation learning, in which the output activation (sum of the squared output) of the neural network estimates the likelihood of the input patterns, or ``learn more, activate more" in simpler terms. For classification on a few small classical datasets, activation learning performs comparably to backpropagation using a fully connected network, and outperforms backpropagation when there are fewer training samples or unpredictable disturbances. Additionally, the same trained network can be used for a variety of tasks, including image generation and completion. Activation learning also achieves state-of-the-art performance on several real-world datasets for anomaly detection. This new learning paradigm, which has the potential to unify supervised, unsupervised, and semi-supervised learning and is reasonably more resistant to adversarial attacks, deserves in-depth investigation.
\end{abstract}

% \begin{IEEEkeywords}
% IEEE, IEEEtran, journal, \LaTeX, paper, template.
% \end{IEEEkeywords}

% \fi
\IEEEpeerreviewmaketitle

\section{Introduction}\label{sec1}

The backpropagation algorithm  \cite{Rumelhart1986} has driven the recent success of machine learning in tasks such as speech and image recognition \cite{Krizhevsky2012}, language processing, image and music creation, playing human games \cite{Silver2016}, etc. Many scientists argue that the backpropagation algorithm, despite being a highly effective tool for training neural networks by minimizing specific loss functions, is different from the rules governing human learning \cite{Crick1989,McClelland2006,Marblesone2016,Whittington2019}. One limitation of backpropagation is that the features learned by minimizing a particular loss function tend to be task-specific. This makes it difficult for the trained models to perform generic tasks, necessitates a large amount of labeled data, and renders them vulnerable to adversarial attacks \cite{Madry2017}. Inspired by the brain, which is believed to learn in a predominantly unsupervised fashion \cite{Grossberg1987, Zaadnoordijk2022}, we intend to create a new learning paradigm that enables forward unsupervised training of neural networks based on a simple local learning rule while achieving comparable performance to backpropagation. The fundamental idea is that, when every neuron in a layer competes to activate while presenting distinct features, the network transmits the maximum amount of information to the next layer, and learning is enforced.

Hebbian plasticity is a local correlation-based learning rule proposed by Hebb that has been supported by experimental evidences such as long-term potentiation and depression \cite{Bear1994, McClelland2006}. It is simply phrased as `cells that fire together wire together.' To train multiple neurons in a layer with Hebbian plasticity, a competitive mechanism of `winner takes all' \cite{Rumelhart1985} was introduced to raise competition among neurons such that only the neuron with the strongest  sum of synaptic input is excited to update weights. However, Hebbian plasticity was long thought to be impractical and inefficient for training artificial neural networks, until it was recently discovered that Hebbian plasticity combined with `winner takes all' can learn lower-layer features to achieve comparable performance to networks trained end-to-end with backpropagation \cite{Krotov2019}. Further efforts concentrated on applying Hebbian plasticity to convolutional networks \cite{Grinberg2019,Gupta2021,Lagani2021,Miconi2021} or multi-layer networks \cite{2019Amato,Shinozaki2020,Journe2022}. However, these methods are not entirely backpropagation-free, i.e., the top layer is still trained via backpropagation for certain tasks such as classification, and the `winner takes all' is also not perfectly local when the neurons in a layer are not connected to each other.

We develop a local competitive learning rule for updating connection weights that imposes competition among neurons without relying on the global `winner takes all.' Specifically, the connection weight $w_{ij}$ between neuron $i$ in a layer and neuron $j$ in the layer above is adjusted by
\begin{equation} \Delta w_{ij} = \eta y_j(x_i - \sum_{k}y_{k}w_{ik}), \end{equation}
where $x_i$ is the input from neuron $i$ to neuron $j$, $y_j=\sum_{u} x_u w_{uj}$ is the weighted sum of the input for neuron $j$, and $\eta$ is the learning rate. In this rule, the term $\sum_{k}y_{k}w_{ik}$ (summing over all neurons that take $x_i$ as input) for synaptic weight decay is the key to raise local competitions. Using this learning rule, a trained layer can decompose an input pattern to some non-orthogonal principal components, and the learnt feature can approximately reconstruct the input data. This reconstruction capability makes learning more resistant to adversarial attacks \cite{Qin2019}. This learning rule can be used to build unsupervised pre-training models to enhance the performance of some supervised learning tasks, especially when there are more unlabeled data. Moreover, this local learning rule enables the development of a new learning paradigm called activation learning, in which the output activation of the neural network estimates the likelihood of input patterns,  or `learn more, activate more' in simpler terms.

Activation learning uses a multi-layer neural network that accepts both data and labels (or purely data) as inputs, where each layer is trained without supervision based on the local competitive learning rule and modulated by a magnitude-preserving activation function. The output activation (the sum of the squared output) of a trained network tends to be upper bounded by the input strength (the sum of the squared input) and can be used as a universal distribution estimator. Due to the task-independent nature of activation learning, the network model is generic and applicable for general purposes. With the same network, a discriminative task is to find the missing category from the data by maximizing the output activation, whereas a generative task aims to infer the missing data from a given category by injecting randomness.

Similar to human learning, activation learning for classification can be enhanced by the feedback of accuracy information, which provides negative samples (data with incorrect labels) for the network to unlearn and modulates the global learning rate. Activation learning achieves comparable performance to backpropagation on MNIST and CIFAR-10 using a fully connected neural network without complicated regularizers and data augmentation. In addition, it exhibits several advantages over backpropagation, including improved performance for fewer training samples, robustness against external disturbances, and resistance to adversarial attacks. Using $600$ labeled training samples from MNIST, activation learning can reach a classification error rate of $9.74\%$, while backpropagation can only get an error rate of $16.17\%$ with one additional output layer.  This is because activation learning is more capable of learning plentiful local features from small pieces of each input pattern. Subsequent research using activation learning for anomaly detection demonstrated that activation learning is not just a new learning theory but also a practical technology that outperforms backpropagation in certain real-world scenarios.

The rest of this paper is organized as follows. Section \ref{section_local_learning} introduces and analyzes the local competitive learning rule, which raises competition among neurons. The framework of activation learning is presented in Section \ref{section_activation_learning}, followed by experiments on MNIST in Section \ref{section_classification}. Section \ref{section_local_connections} examines activation learning with local connections on CIFAR-10. Section \ref{section_discussion} discusses several further practical or prospective benefits of activation learning as well as its relevance to the Forward-Forward algorithm, before drawing conclusions.

\section{Local Competitive Learning Rule}
\label{section_local_learning}

\subsection{Overview of Hebbian Learning}

Hebb's proposal for learning is to modify the connection weights of neurons in accordance with the principle `cells that fire together wire together,' which has inspired a multitude of studies on learning systems and synaptic plasticity. The connection weight $w_{ij}$ from a neuron $i$ in a layer to neuron $j$ in the layer above increases if both neurons fire concurrently. Formally, the original version of Hebbian plasticity is
\begin{equation}\Delta w_{ij} = \eta x_iy_j\end{equation}
with $x_i$ serving as the $i$th input to neuron $j$, $y_i$ being the output of neuron $j$, and $\eta$ being the learning rate. This version, however, is not stable for learning, as all weights might grow unboundedly until they reach their maximum allowable values.

In a biological neuron \cite{Bourgeois1989, Pallas1991}, when some synapses to a specific postsynaptic neuron are strengthened, other synapses are weakened \cite{Guillery1972, Miller1996}. To ensure convergence, constraints are added to various Hebbian plasticity models \cite{Miller1994}, such as limiting the synaptic strength sum of a neuron. Such constraints can be implemented locally via mechanisms such as spike-timing-dependent synaptic plasticity \cite{Song2000}, activity-dependent synaptic scaling \cite{Turrigiano1998, Davis1998}, and the sliding threshold of the BCM model \cite{Bienenstock1982}. One variant of Hebbian plasticity proposed by Oja \cite{Oja1982} is
\begin{equation}\Delta w_{ij} = \eta y_j(x_i - y_jw_{ij}),\label{equ1}\end{equation}
in which $y_j=\sum_{i}x_{i}w_{ij}$ is the weighted sum of the input. This rule, which enforces a constraint on the total synaptic weight of a neuron via synaptic weight decay, is effective for training a neuron. However, Oja's rule only works for training a single neuron; for multiple neurons to learn different features, competitions are required. Accumulating evidence reveals the crucial role of competitions in shaping learning activity \cite{Scanziani2011, Markram2004}. In the cerebral cortex, a class of neurons called inhibitory neurons constitutes around $20\%$ of the cortical neuronal population \cite{Meinecke1987}. When inhibitory neurons are active, they release the transmitter GABA, which inhibits the firing of other neurons.

Sparsity was exploited \cite{Rumelhart1985, Rosenblatt, Malsburg1973, Crossberg, Fukushima1975} to model and raise competition among neurons in a layer.  For example, relying on the assumption that neighboring neurons tend to excite each other and neurons at a greater distance tend to inhibit each other \cite{Kohonen1982, Linsker1986} in a two-dimensional neuron layer, sparsity is used to model refinement of map-like cortical representation \cite{Goodhill1993}. A popular competition mechanism is
`winner takes all' \cite{Rumelhart1985}, which divides the neurons in a layer into a group of inhibitory clusters, in which only the neuron with the strongest sum of synaptic input in each cluster is excited, thus inhabiting other neurons in that cluster. However, the `winner takes all' mechanism disrupts the locality of plasticity when the neurons in a layer are not fully connected. The competition mechanism of `winner takes all' combined with biologically plausible synaptic modification, has recently attracted increasing attention for training lower-layer features without supervision \cite{Krotov2019, Grinberg2019, Gupta2021,Lagani2021,Miconi2021, 2019Amato,Shinozaki2020,Journe2022}. Based on the learned unsupervised features, a linear classifier can achieve performance comparable to the entire network trained end-to-end by backpropagation, indicating that the potential of Hebbian learning has been underestimated. Note that these existing biologically plausible methods require backpropagation to train the top-layer classifier, i.e., they are not entirely backpropagation-free.

\subsection{Local Competitive Learning}

\begin{figure}[!t]
\centering
\includegraphics[width=5.6in]{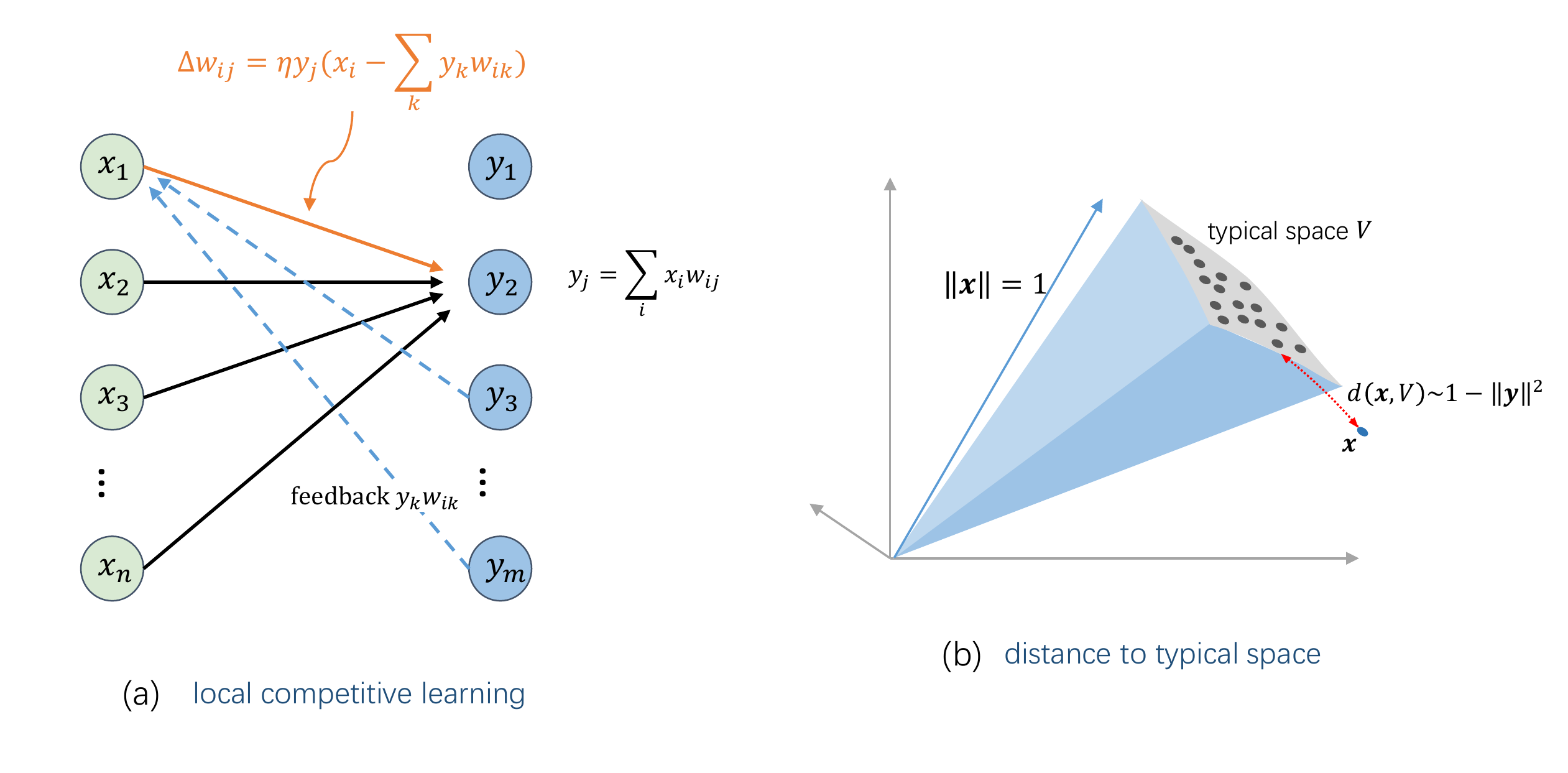}
\caption{(a) The local competitive learning rule. (b) A geometric interpretation of the local competitive learning rule.}
\label{fig_feature_learning}
\end{figure}

We develop a local competitive learning rule for connection weight modification that enforces competition among neurons without employing `winner takes all,' where the modification of each neuron only depends on its activity or the activity of its neighbors.
Specifically, the modification of the connection weight $w_{ij}$ between a neuron $i$ and a neuron $j$ in the layer above is given by
\begin{equation}
\Delta w_{ij} = \eta y_j(x_i - \sum_{k}y_{k}w_{ik}), \label{equ2}
\end{equation}
with the input $x_i$ and the net input $y_j=\sum_{u}x_{u}w_{uj}$. In this rule, the input $x_i$ is adjusted by internal feedback from all the receiving neurons, i.e., $\sum_{k}y_{k}w_{ik}$, to raise competition among neurons, as illustrated in Fig. \ref{fig_feature_learning}(a). This feedback term is made of two components: the feedback $y_jw_{ij}$ from neuron $j$ and the feedback $\sum_{k\neq j}y_{k}w_{ik}$ from the other receiving neurons. The feedback from neuron $j$, which was previously studied in Oja's rule (\ref{equ1}), serves to limit the total synaptic weights over neuron $j$. The feedback from the other receiving neurons forces them to compete with neuron $j$ to fire, enabling neurons in the same layer to represent different features.  To see this, if another receiving neuron $k\neq j$ is excited, the connection weight $w_{ij}$ to neuron $j$ is weakened by a magnitude of $\eta y_j y_{k} w_{ik}$ (assuming all the variables are positive) contributed by neuron $k$'s excitation. As a result, concurrently activated neurons inhabit each other, causing differential connection weights for individual neurons.

After a sufficient number of training steps with a very small learning rate $\eta$, the connection weights $\mathbf{w}$ tend to converge to asymptotically stable solutions of
\begin{equation}
 C \mathbf{w} - \mathbf{w} \mathbf{w}^T C \mathbf{w} = 0, \label{equ4_4}
\end{equation}
where $C=E\{\mathbf{x} \mathbf{x}^T\}$ is the covariance matrix of the training samples. There are two stable solutions of $\mathbf{w}$ corresponding to $\mathbf{v}$ and $-\mathbf{v}$ when there is a single neuron, with $\mathbf{v}$ being the eigenvector corresponding to the largest eigenvalue of the covariance matrix $C$. When the number of receiving neurons exceeds two, there are an infinite number of stable solutions. Which stable solution the learning rule converges to depends on the initial value of $\mathbf{w}$
and the training samples.

\subsection{Mathematical Properties}

Several results can be derived from the condition of stable solutions that aid in the comprehension of the proposed learning rule's functioning mechanisms. Appendix \ref{Appendix_rule} provides detailed mathematical proofs of the results and more.

\textbf{Property 1 (Reconstruction)}: The local competitive learning rule performs best towards minimizing the mean reconstruction error \begin{equation}E\|\mathbf{x} - \mathbf{wy}\|^2,\label{equ2_1}\end{equation}
where $\mathbf{x}$ is the input vector and $\mathbf{y}=\mathbf{w}^T \mathbf{x}$; hence, the output $\mathbf{y}$ of a trained layer can approximately reconstruct the input data $\mathbf{x}$ by $\mathbf{wy}$. This capacity for layer reconstruction makes learning more resistant to adversarial attacks \cite{Qin2019}.

\textbf{Property 2 (Non-orthogonal PCA)}: The mean reconstruction error of the local competitive learning rule converges to that of the principal component analysis (PCA) \cite{Chan2015}, but unlike PCA, the components are typically non-orthogonal. This ensures that as much information as possible is passed to the next layer while reducing the network's vulnerability to the failure of a few neurons corresponding to the principal components.

\textbf{Property 3 (Convergence)}:  There are infinitely many stable solutions when the number of receiving neurons $m>1$.  The sum of squared synaptic weights $\|\mathbf{w}\|= \sum_{ij} w_{ij}^2$ tends to converge to the number of neurons $m$ despite the absence of an explicit constraint or normalization in the learning rule.

\textbf{Property 4 (Activation)}:  The output activation $\|\mathbf{y}\|^2=\|\mathbf{w}^T\mathbf{x}\|^2$ tends to be upper bounded by the input strength $\|\mathbf{x}\|^2$.  If the input strength $\|\mathbf{x}\|^2$ is fixed, $\|\mathbf{y}\|^2$ reflects how well the input $\mathbf{x}$ can be reconstructed. Hence the output activation can be used to estimate the typicality of the input pattern.

These properties show that the local competitive learning rule can approximately reconstruct the input from the output, while the output activation can be used as a measurement of the input's typicality. A geometric interpretation is presented in Fig. \ref{fig_feature_learning}(b), where the typical space is a convex and compact area in the high-dimensional sphere that contains majority of the training samples, and $1-\|\mathbf{y}\|^2$ roughly indicates how close the input pattern $\mathbf{x}$ is to the typical space when $\mathbf{x}$ is normalized. By cascading the local competitive learning rule with multiple layers, the shape of the typical space can be refined further to approximate complicated distributions of the input samples. This motivates the emergence of activation learning.

\subsection{Forward Unsupervised Learning}

\begin{figure}[!t]
\centering
\includegraphics[width=5.2in]{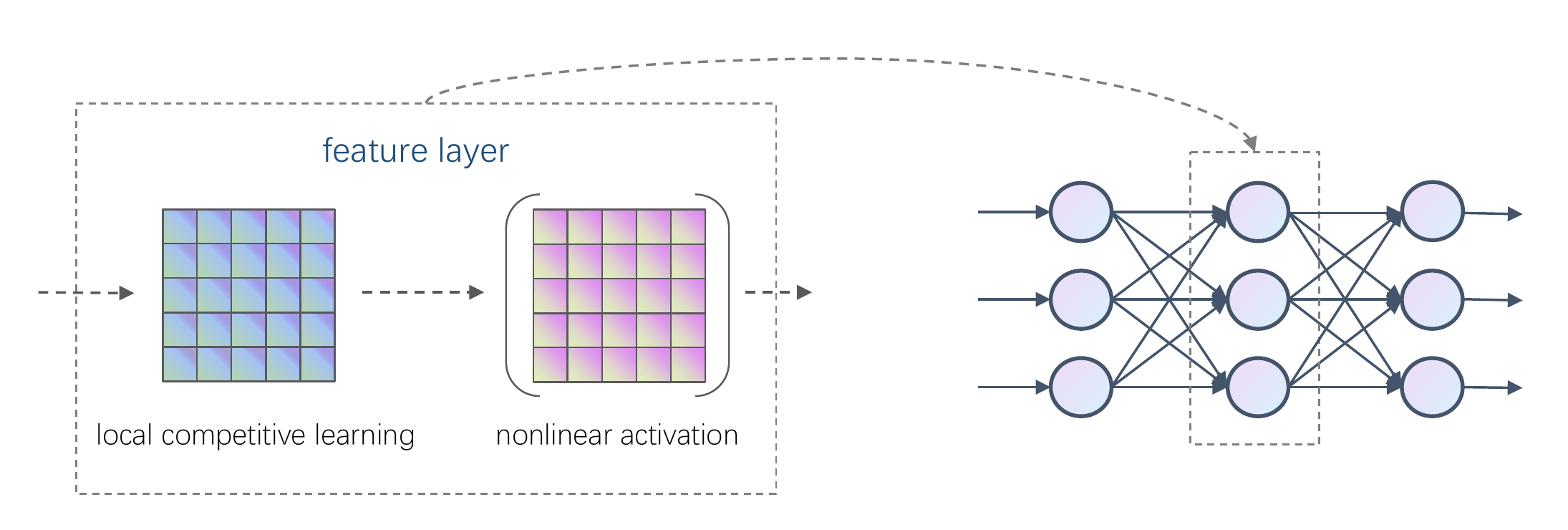}
\caption{A multi-layer network for bottom-up unsupervised learning in which each layer is composed of a linear transformation trained by the local competitive learning rule and a nonlinear activation function.}
\label{fig_feature_learning_network}
\end{figure}

The local competitive learning rule is inherently applicable to bottom-up unsupervised neural network training. Given a multi-layer neural network, as depicted in Fig. \ref{fig_feature_learning_network}, each layer trained by the local competitive learning rule extracts features automatically from its input, which is then regulated by an activation function. The network can be trained with all layers together for each batch of training samples or layer by layer, i.e., training a layer commences after the lower layers have been trained and frozen. All the layers form a chain for unsupervised feature extraction, with the lower layers representing low-level features and the higher layers representing high-level features.

\begin{figure}[!t]
\centering
\includegraphics[width=5.4in]{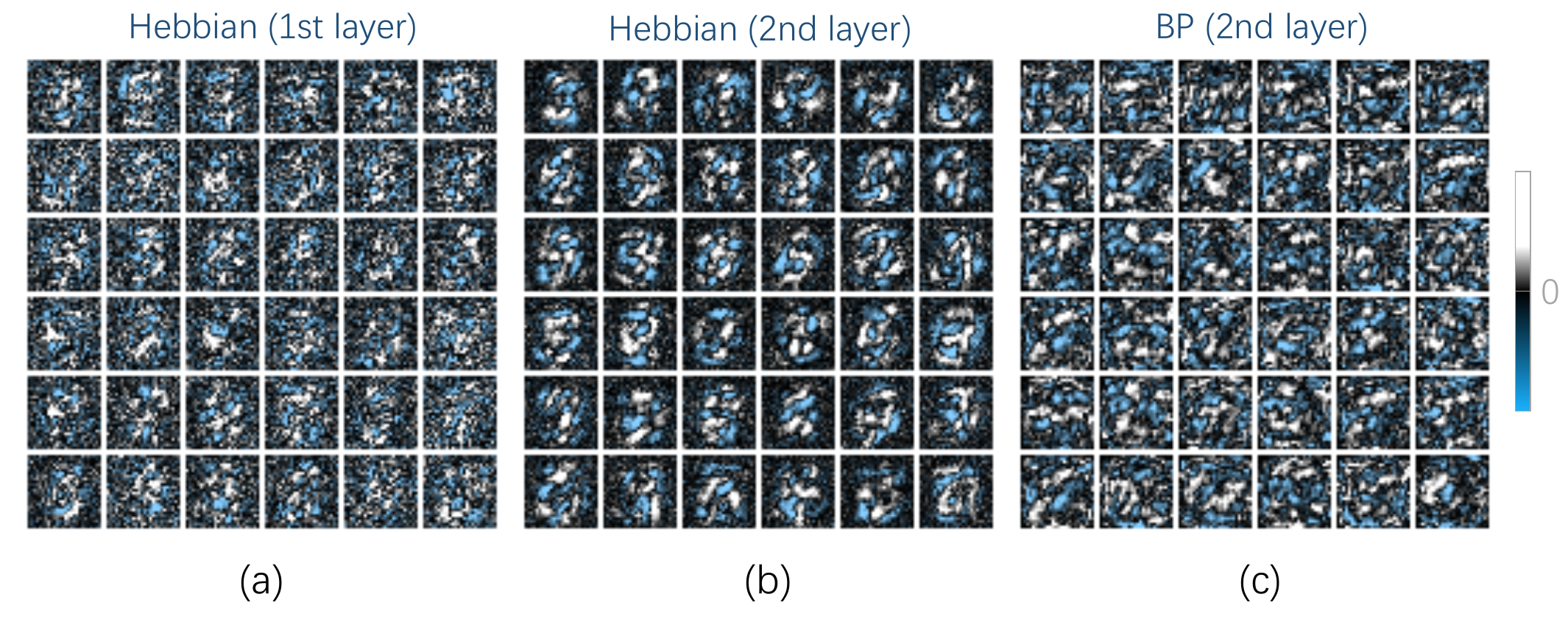}
\caption{Learned features from MNIST by the local competitive learning rule and backpropagation (BP) respectively. (a) The first layer features of the $36$ most excited units learned by the competitive learning rule. (b) The second layer  features learned by the competitive learning rule. (c) The second layer features learned by backpropagation.}
\label{fig_network_feature}
\end{figure}

Fig. \ref{fig_network_feature} plots the unsupervised features learned by the local competitive learning rule using a two-layer neural network from MNIST (a database of black-and-white handwritten digit images with $28\times 28$ pixels each), and compares them with those learned by backpropagation using the same neural network connected to $10$ output units at the top. Each network layer consists of $28\times 28$ units. The plotted features are generated by calculating the gradient of the target unit with respect to the input image. White represents positive values, while blue represents negative values, reflecting whether the input pixel excites or inhabits the target neuron unit. Consistent with the fact that pixels at the image's periphery are often uninformative, the features learned by the local rule are near zero at the periphery of the images. However, the features learned by backpropagation appear more random and dispersed across the entire image, which makes the network more susceptible to adversarial attacks \cite{Krotov2019}.

The local competitive learning rule is more capable than backpropagation of discovering abundant features from a few shots of samples. As studied in Appendix \ref{section_pretraining}, the local competitive learning rule can be used to construct unsupervised pre-training models for improving the performance of certain learning tasks or reducing the amount of labeled data required. It also enables the development of activation learning as a new learning paradigm.

\section{Activation Learning}
\label{section_activation_learning}

During the process of learning a digit, the brain receives both visual and auditory signals about the digit. It is doubtful that the brain will intentionally choose the auditory signal as the label and the visual signal as the data. It may instead accept both signals as input and learn their association in an unsupervised manner, allowing retrieval of the auditory signal from a visual input and vice versa. This motivates activation learning, which attempts to discover correlations within all input patterns based on the local competitive learning rule and whose output activation estimates the likelihood of the input pattern.

\subsection{Paradigm of Activation Learning}

Activation learning is based on a simple multi-layer neural network, as illustrated in Fig. \ref{fig_feature_learning_network}, in which each layer is trained by the local competitive learning rule without supervision, and the inference is to retrieve the missing units of the input pattern to maximize the output activation (the sum of the squared output) or by sampling. It is built on Property 4 of the local competitive learning rule, which states that a layer's output tends to be more active for more typical inputs. Given that the activation of a single layer can only represent simple distributions, activation learning applies multiple layers to discover more complex distributions. To ensure the transferability of activation strength across layers, activation learning requires (1) that each input sample be normalized and (2) that the activation function does not change the activation strength; that is, the activation function must be a nonlinear function $f$ such that $\|f(\mathbf{y})\|=\|\mathbf{y}\|$ for any $\mathbf{y}$. This allows for the cascading of multiple layers, so that the network output activation tends to decrease as the network grows deeper.

We study several activation function choices that do not alter the input's magnitude. The simplest is the abs (absolute value) function $\lvert \cdot \rvert$.
By folding all the feature vectors into positive ones with this activation function, some clusters of feature vectors may become less dispersed and concentrate in a smaller area. As the local competitive learning rule performs similarly to PCA, which is more effective with zero-mean data, it motivates the selection of an alternative activation function:
\begin{equation}\mathrm{std\_abs}(\mathbf{y}) = \beta  \cdot \mathrm{std} (\lvert \mathbf{y}\rvert),\end{equation}
where $\mathrm{std} (\lvert \mathbf{y}\rvert)$ standardizes $\lvert \mathbf{y}\rvert$ by subtracting the mean value from $\lvert \mathbf{y}\rvert$ and then dividing the result by the standard deviation, and the scalar $\beta$ rescales the standardized data so that the activation function is magnitude-preserving. When the network is trained layer-by-layer, the mean and variance of $\lvert \mathbf{y}\rvert$ can be computed explicitly, leading to its standardization.  When all layers are trained concurrently, standardization can be approximated with batch normalization. Layer normalization that subtracts the layer mean of $\lvert \mathbf{y}\rvert$ from $\lvert \mathbf{y}\rvert$ can achieve comparable performance while being simpler to implement.

Activation learning can be viewed as a universal distribution estimator that explicitly predicts the probability distribution of input data. The probability of an input pattern
$\mathbf{x}$ can be estimated by
\begin{equation} p_\textrm{A}(\mathbf{x}) \propto e^{a\|\mathbf{y}\|^2},\end{equation}
where $\|\mathbf{y}\|^2$ is the network output activation and $a$ is a positive constant. A network trained by activation learning can serve as a generic model since it attempts to learn the statistical distribution of the training samples regardless of the specific tasks. In contrast, the loss function to be minimized in backpropagation is typically task-related.

\begin{figure}[!t]
\centering
\includegraphics[width=5.2in]{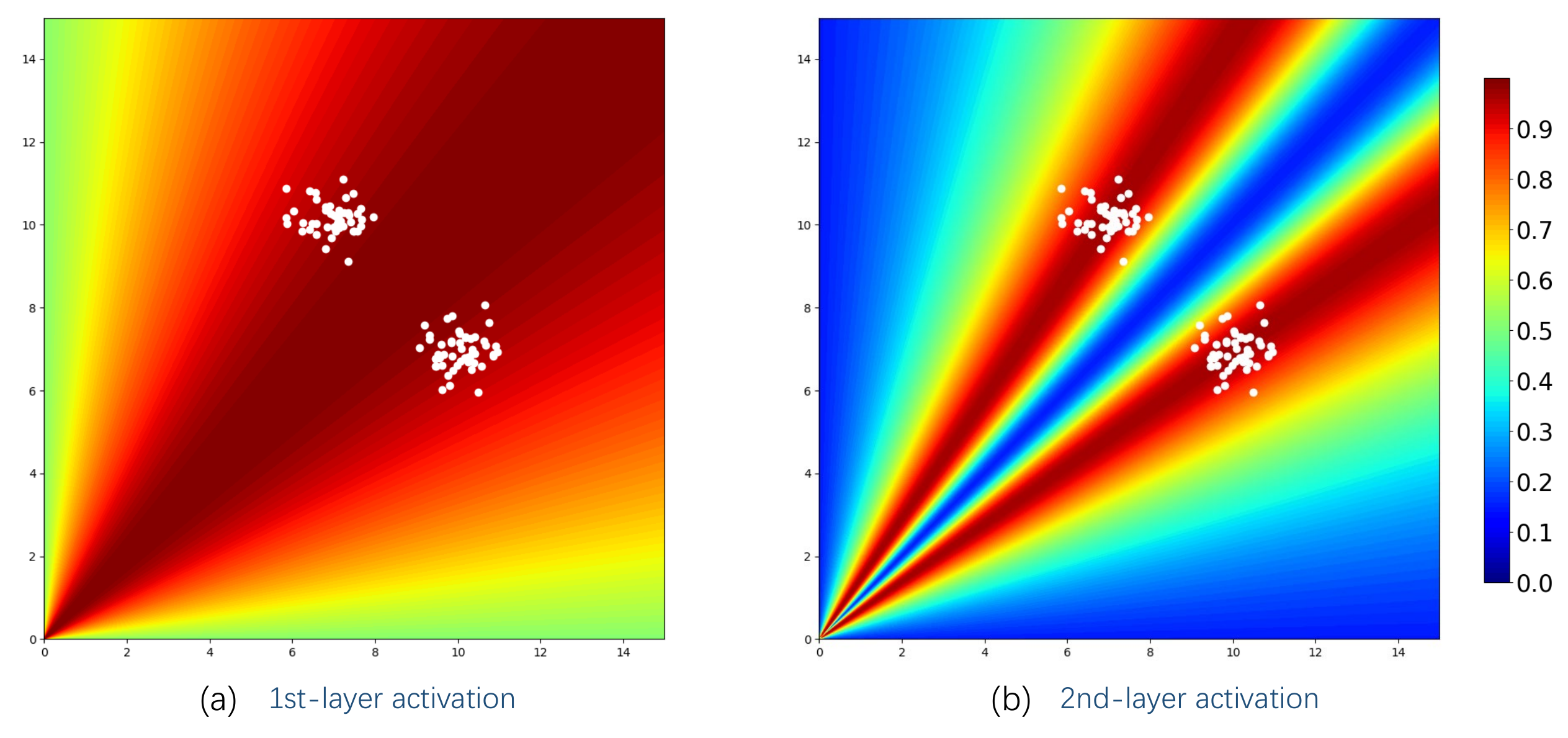}
\caption{The activation of the first and second layers of a two-layer neural network with two units per layer, using the white points as training samples for activation learning.}
\label{fig_gauss_toy}
\end{figure}

A multi-layer network of activation learning can be described as a series of connected filters, with each layer functioning as an activation filter that prunes the shape of the predicted distribution. The output activation of an $l$-layer network is given by
\begin{equation}\|\mathbf{y}\|^2 = \|\mathbf{x}\|^2 h_1(\mathbf{x}) h_2(\mathbf{x}) ... h_l(\mathbf{x})\end{equation}
where $h_i(\mathbf{x})$, often ranging between $[0, 1]$, is the activation filter for the $i$th layer. Fig. \ref{fig_gauss_toy}  is a simple demonstration of activation learning utilizing a two-layer neural network with two units per layer to learn from randomly generated points. It uses the std\_abs activation function. Since the input is normalized, the output activation is dependent only on its angle and not its magnitude. It is evident that the estimated distribution of the first layer mixes all the points, however the pruning of the second layer helps to distinguish them as two separate clusters.

\begin{figure}[!t]
\centering
\includegraphics[width=5.0in]{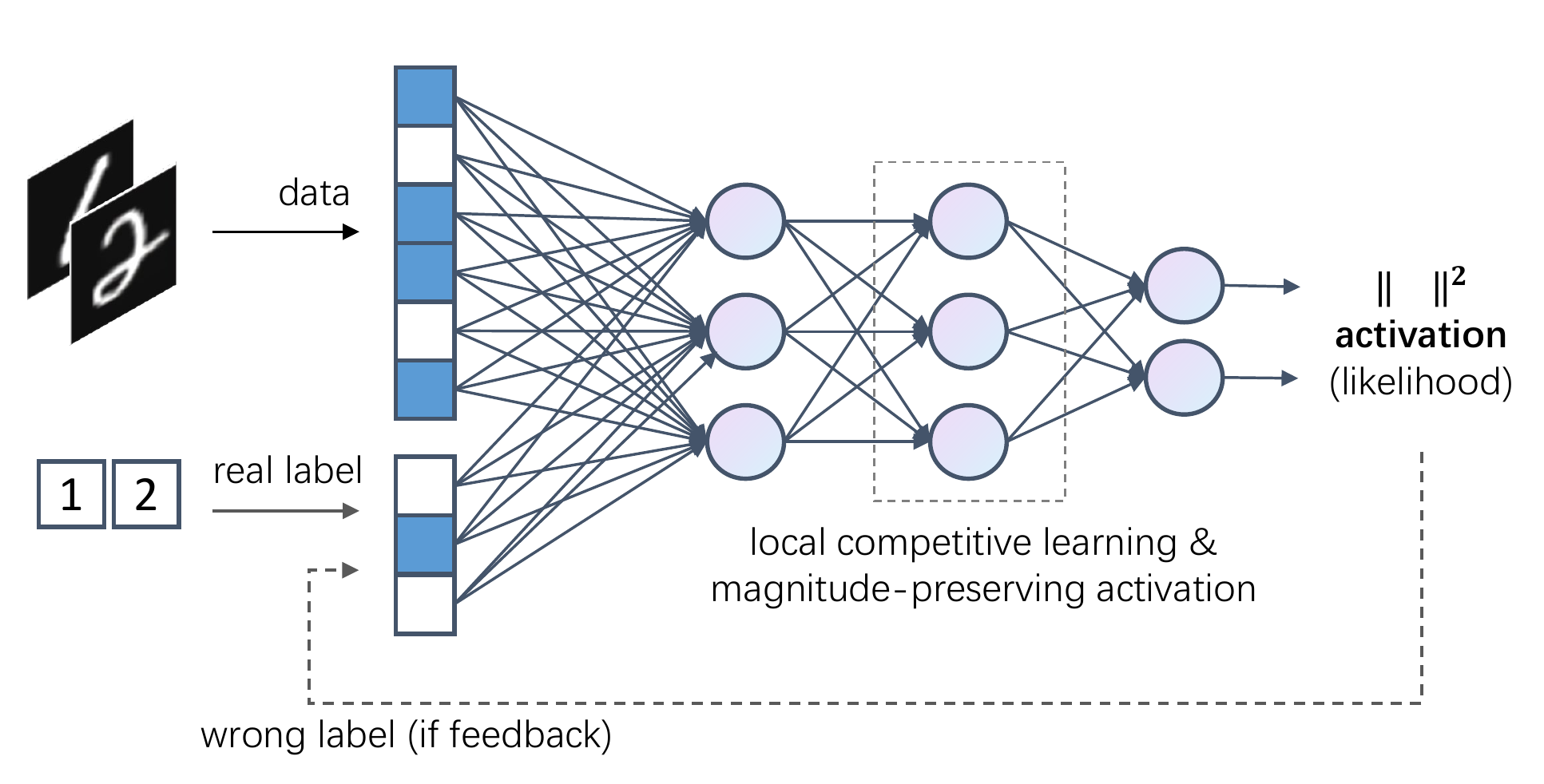}
\caption{An illustration of activation learning for a multi-layer neural network that takes both data and labels as input.}
\label{fig_activation_learning}
\end{figure}

A network of activation learning can receive both data and labels as input, and process them through multiple layers trained by the local competitive learning rule. Typically, input data with the correct label results in a higher output activation than input data with the incorrect label. It differs from a standard neural network, which uses the data as input and the label to compute an output error. Classification in activation learning to identify the label that best combines with the input data to maximize output activation. To improve classification performance, we use data with correct labels as positive samples and data with wrong labels generated based on the feedback of accuracy information as negative samples. The network learns from positive samples at a positive rate and unlearns from negative samples at a negative rate. On the other hand, the same neural network that is used for classification can also be utilized for data generation - to infer missing data based on given categories. As the learning rule of activation learning seeks to explore inherent correlations of the input patterns regardless of specific tasks, the learned models are applicable to a broad variety of learning tasks.

\subsection{Optimization based Inference}

In activation learning, inference is used to retrieve any missing units (or any subset of missing units) from the known ones of the input by maximizing the network's output activation or by sampling. Image generation and classification are examples of inference. Let $\mathbf{a}$ represent the known units of an input pattern, $\mathbf{z}$ represent the missing units and $\mathbf{y}(\mathbf{a}, \mathbf{c})$ serve as the output activation. For maximizing the output activation, the objective is to solve an optimization problem:
\begin{equation}
\mathbf{z}^* = \arg\max_{\mathbf{z}} \|\mathbf{y}(\mathbf{a}, \mathbf{z})\|^2 \label{equ_generative2}
\end{equation}
such that $(\mathbf{a}, \mathbf{z})\in S$, where $S$ is the set of normalized feasible input patterns.

The simplest method for determining the optimal  $\mathbf{z}^*$ is to enumerate all possible $\mathbf{z}$ and choose the best one. This method is suitable for classification with a small number of classes, where $\mathbf{a}$ is the input data and $\mathbf{z}$ is the class's encoded representation.

Another method is to find a locally optimal solution using the gradient descent method. For data generation, $\mathbf{a}$ is the given encoded  representation of the class, and $\mathbf{z}$ is for the normalized data pattern to infer such that $\|\mathbf{z}\| = 1$. The problem (\ref{equ_generative2}) becomes a constrained non-convex problem. It can be converted into an unconstrained problem:
\begin{equation}
\mathbf{z}^* = \arg\max_{\mathbf{z}} \|\mathbf{y}(\mathbf{a}, \frac{\mathbf{z}}{\|\mathbf{z}\|})\|^2 - (\|\mathbf{z}\|^2-1)^2, \label{equ_alternative}
\end{equation}
in which the term $(\|\mathbf{z}\|^2-1)^2$ forces $\|\mathbf{z}\|^2$ to be very close to $1$.  The gradient decent method can be utilized to find a locally optimal solution for this equivalent problem.

%\subsection{Propagation based Inference}
%
%\begin{figure}[!t]
%\centering
%\includegraphics[width=3.6in]{figures/inference_propagation}
%\caption{An illustration of propagation for inference.}
%\label{fig_inference_propagation}
%\end{figure}
%
%
%Activation learning has a nice property that each layer's output can reconstruct its input data, contributed by the local competitive learning rule. This enables propagation over the network to discover the missing units of input. A rough idea is that the given units of input can be used to approximately calculate the higher-layer features and the network output, which can further reconstruct the whole input pattern. Hence, we present a propagation based inference method on a multi-layer neural network for activation learning, as the one used for Hopfield network. Fig. \ref{fig_inference_propagation} is a simple illustration of the inference method.

% \subsection{Propagation based Inference}
% If the activation function is a monotonic function, then it is easier to derive propagation-style inference.
% 1. End-to-End Propagation Algorithm.
% 2. Local Iterative Propagation Algorithm.
% Iteration: Up * $\alpha$ + Down * (1-$\alpha$)

\subsection{Classification with Feedback}

It has been observed that the feedback of accuracy information plays an important role in enhancing human learning \cite{McCandliss2002,McClelland2006}, by providing something beyond the simple correlation-based strengthening of synaptic connections. This motivates us to add accuracy feedback into activation learning for improving classification accuracy. The accuracy feedback acts as a teacher who corrects a model's mistakes. When the model is prone to making errors on input data, it instructs the model to learn the data with the correct label at a positive learning rate and unlearn the data with the incorrectly classified label at a negative learning rate.

In order to correctly classify data $\mathbf{a}$, activation learning anticipates a suitably large activation gap
\begin{equation}g(\mathbf{a})= \|\mathbf{y}(\mathbf{a}, c)\|^2- \|\mathbf{y}(\mathbf{a}, c')\|^2,\end{equation}
where $c$ is the correct label and $c'\neq c$ is the incorrect label that produces the strongest output activation. The accuracy feedback not only stimulates the positive sample $(\mathbf{a}, c)$, raising $\|\mathbf{y}(\mathbf{a}, c)\|^2$, but also inhabits the negative sample $(\mathbf{a}, c')$, decreasing $\|\mathbf{y}(\mathbf{a}, c')\|^2$. Specifically, let $\Delta w_{ij}^+$ represent the change of the connection weights when the local competitive learning rule is applied to the positive sample $(\mathbf{a}, c)$,
and let $\Delta w_{ij}^-$ represent the corresponding change of the negative sample $(\mathbf{a}, c')$. With feedback, the overall change in the connection weights is
\begin{equation}\Delta w_{ij} = \gamma (\Delta w_{ij}^+ - \lambda \Delta w_{ij}^-),\label{equ5_3}\end{equation}
where $\gamma\geq 0$ is a non-increasing function of the activation gap $g(\mathbf{a})$ that modulates the global learning rate, and $\lambda\in [0, 1]$  determines the weight of unlearning negative samples, also known as the unlearning factor.  This learning rule results in a higher learning rate for incorrectly classified samples than correctly classified ones, or `learn more on errors.' The unlearning factor $\lambda$ is expected to be close to $1$, but not too close, as the learning may not converge.

\section{Experiments with MNIST}
\label{section_classification}

\begin{figure}[!t]
\centering
\includegraphics[width=5.2in]{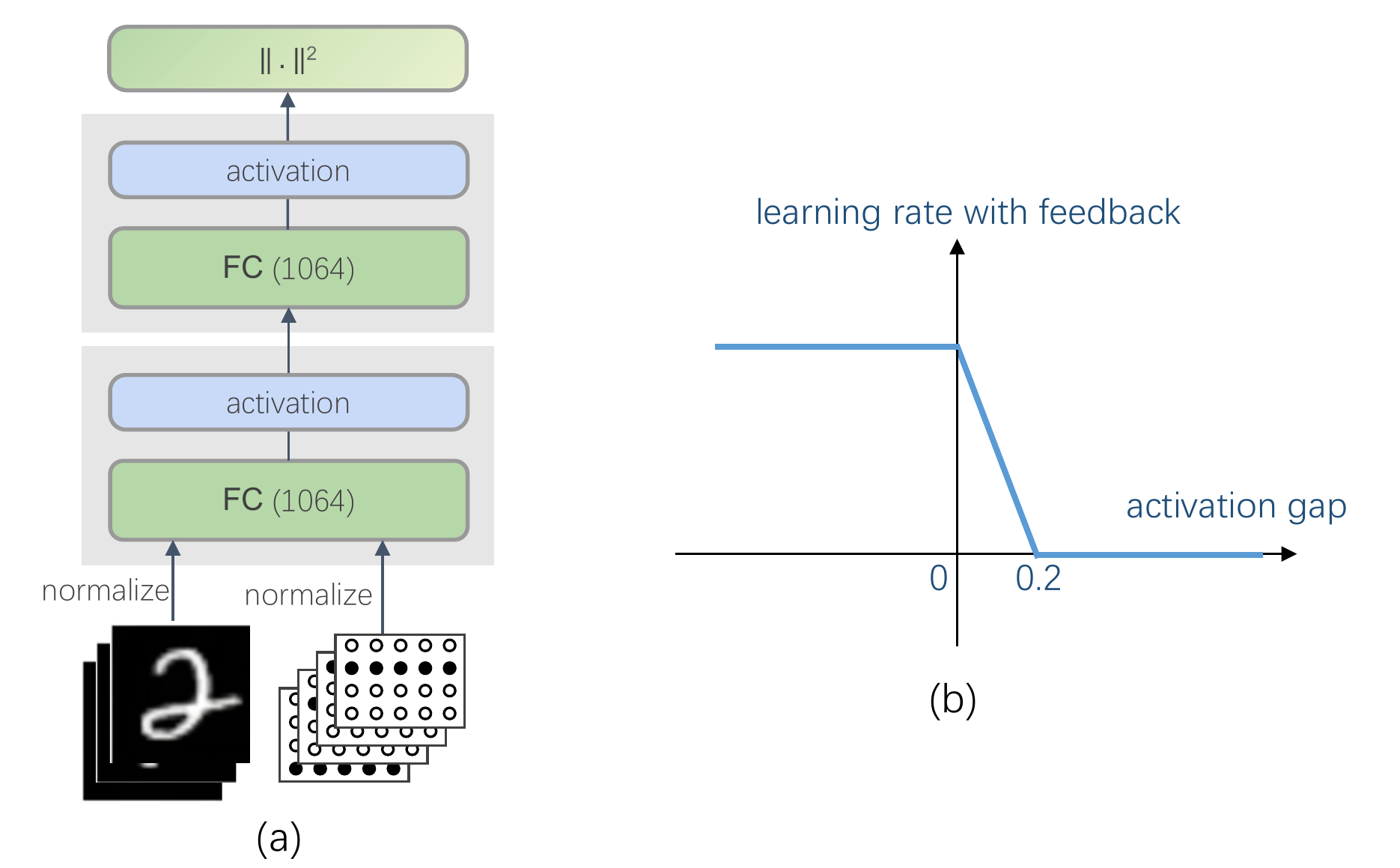}
\caption{(a) The network structure for activation learning on MNIST, which consists of $h$ layers with $1064$ units (the same dimension as the input) in each layer trained by the local competitive learning rule. (b) A truncated linear unit that modulates the global learning rate based on the activation gap between positive and negative samples.}
\label{fig_activation_network}
\end{figure}

This section conducts experiments with the MNIST dataset to investigate the performance of activation learning.
In the experiments, a simple fully connected network structure as depicted in Fig. \ref{fig_activation_network}(a) is used, which accepts as inputs both the normalized $28\times 28$  digit images and the normalized $10\times 28$ encoded representations of their labels. The encoded representation of digit $i$ is a $10\times 28$ matrix with all ones in the $i$th row and all zeros in the other rows.  This network is composed of multiple layers with $1064$ units in each layer (the same size as the input) and connection weights initialized from a normal distribution with a zero mean and small variance. All the layers are trained simultaneously with a batch size of $100$.

\subsection{Learning without Feedback}

\begin{figure}[!t]
\centering
\includegraphics[width=6.0in]{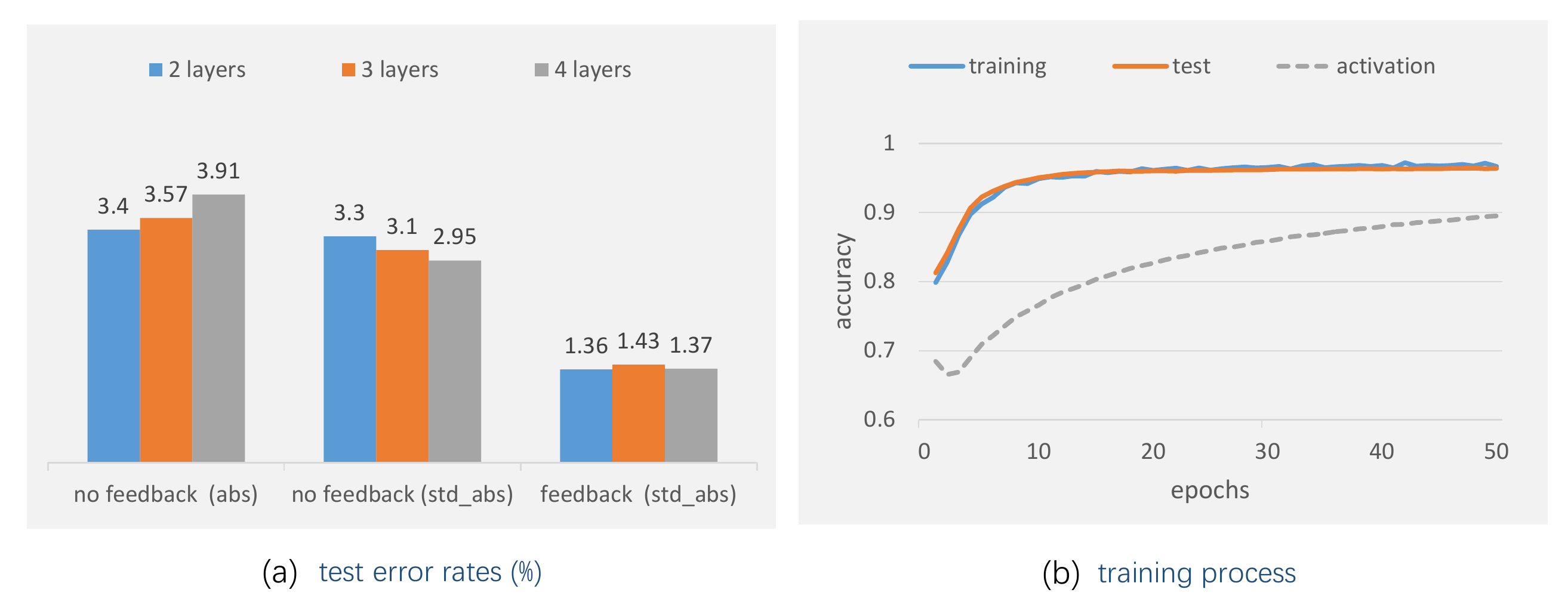}
\caption{(a) The test classification error rates of activation learning with or without accuracy feedback, when the network has two, three, or four layers, respectively. (b) The training/test accuracy and the normalized output activation as training progresses for a two-layer network using the std\_abs activation function.}
\label{activation_classification}
\end{figure}

We investigate activation learning for classification using all $60000$ labeled training images, of which $5000$ are for validation. When classifying a digit image, it searches for the missing label representation that  combines with the provided input image to maximize output activation. Without the feedback of accuracy information, a single-layer network (no hidden layers) can reach a test error rate of approximately $13.6\%$ when early stopping is applied, where the test accuracy first climbs and then declines monotonically as training advances. As the number of network layers increases from one to two, the error rate decreases to about $3.30\%$, see Fig. \ref{activation_classification}(a). In this case, the performance of using abs as the activation function is comparable to that of std\_abs. When a network has more than two layers, the std\_abs activation function enables the addition of more layers for improved performance. With four layers, the error rate is reduced to approximately $2.95\%$, which is close to human  accuracy (the error rate is around $2\%$-$2.5\%$ \cite{Simard1992}). In contrast, as the network depth increases, the error rate using the abs activation function increases.

Fig. \ref{activation_classification}(b) depicts the output activation and classification accuracy of a two-layer network as training advances. The activation of the output tends to increase monotonically and converge to a maximum value bounded by the input strength. In the given number of training epochs, the training accuracy is very close to the corresponding test accuracy, and both tend to rise monotonically and converge quicker than the output activation.

\subsection{Learning with Feedback}

Neural networks with a few fully connected layers trained by backpropagation without using complicated regularizers such as dropout typically reach about a $1.4\%$ test error rate, establishing a baseline.  As a comparison, activation learning with feedback can get about $1.36\%$ test error with a two-layer neural network, as shown in Fig. \ref{activation_classification}(a), where the feedback of accuracy information plays a crucial role. However, performance does not increase as network depth grows.

For learning with feedback, each training sample with the correct class is treated as a positive sample, whereas the image with the incorrect class that maximizes output activation is treated as a negative sample. Let $\Delta w_{ij}^+$ represent the modification of the connection weights on the positive sample based on the local competitive learning rule, and $\Delta w_{ij}^-$ represent the modification of the connection weights on the negative sample. The rule for learning with feedback is to learn the positive sample and unlearn the negative one:
\begin{equation}\Delta w_{ij} = \gamma (\Delta w_{ij}^+ - \lambda \Delta w_{ij}^-).\label{equ_mnist_feedback}\end{equation}
In the experiments on MNIST, we select $\lambda = 0.9$ and $\gamma = \min(\max(5\cdot g+1,0),1)$, a truncated linear function of the activation gap $g$ between the positive sample and negative sample, illustrated in Fig. \ref{fig_activation_network}(b). Note that a truncated linear function may not be the optimal choice to control the global learning rate; other options include exponential or sigmoid functions, where the objective is to learn more on samples that are more prone to have classification errors.

\subsection{Learning from Fewer Samples}

\begin{figure}[!t]
\centering
\includegraphics[width=6.0in]{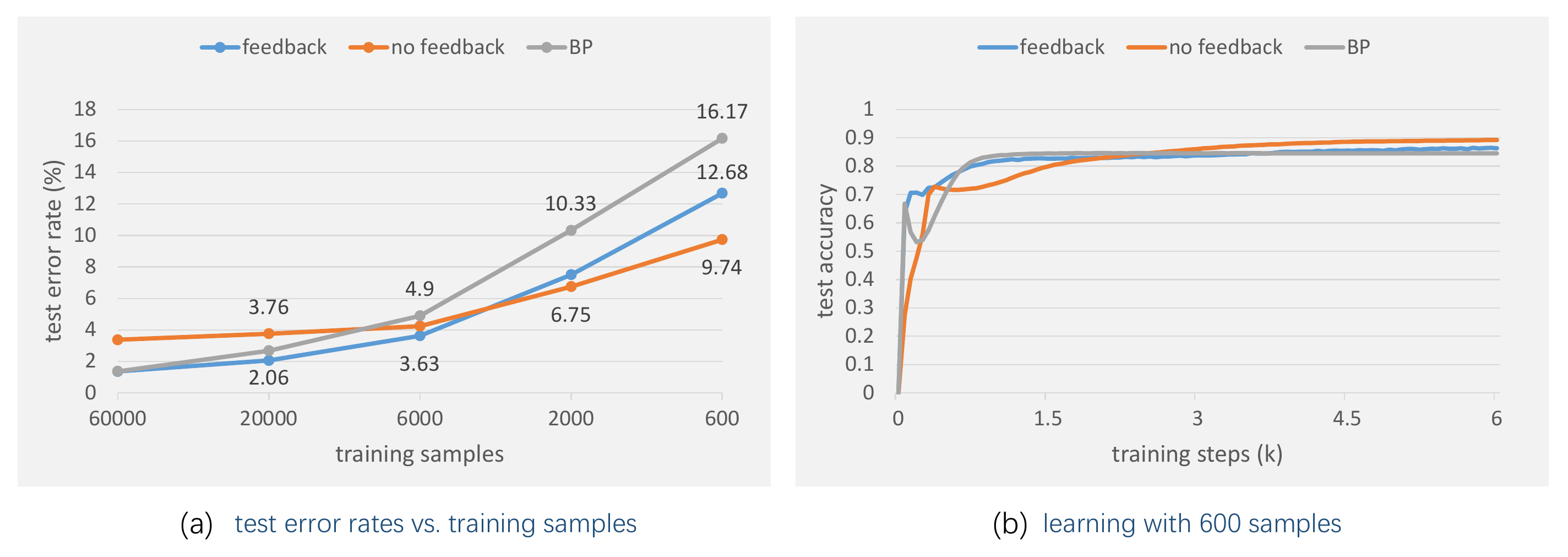}
\caption{(a) The test error rates of a two-layer network trained on a specific number of labeled samples using activation learning, compared to the test error rates of a three-layer network (two hidden layers with $10$ more units on top) trained by backpropagation. (b) The test accuracies as training progresses based on $600$ training samples. }
\label{fig_sample_complexity}
\end{figure}

Activation learning is exhibiting outstanding learning ability with a relatively small number of training samples. Fig. \ref{fig_sample_complexity}(a) studies how the number of training samples affects the error rates of activation learning with a two-layer network, and compares them with the error rates of a three-layer network (two hidden layers with $10$ more units on top) trained end-to-end by backpropagation. Each hidden layer of the three-layer neural network trained by propagation consists of $1064$ units (the same number as the two-layer network for activation learning) with batch normalization and ReLU activation. In the experiments, the test accuracies of activation learning and backpropagation do not decrease as training advances; witness the example of $600$ training samples in Fig. \ref{fig_sample_complexity}(b).  Consequently, none of the training samples are used for validation, and training ceases at $100$ epochs. The experimental results demonstrate that activation learning with feedback outperforms backpropagation for any given number of training samples, despite the network trained by backpropagation having one additional layer. When the number of training samples is less than about $4000$, activation learning without feedback performs the best, indicating that feedback may weaken the model's generalization ability. When the number of training samples is decreased to $600$, the test error rate of activation learning without feedback can reach approximately $9.74\%$, whereas the error rate of a network trained by backpropagation can reach as high as $16.17\%$. Activation learning appears to be efficient for learning from relatively few training samples.

\begin{figure}[!t]
\centering
\includegraphics[width=6.0in]{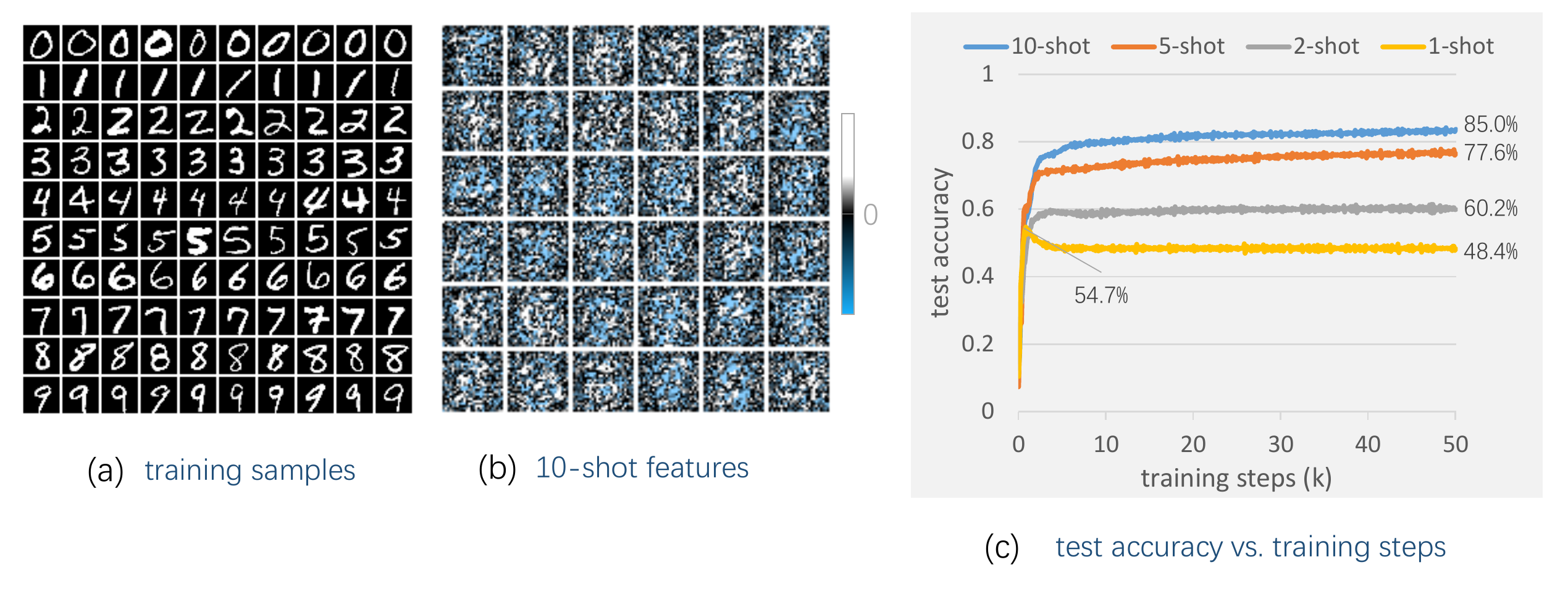}
\caption{Few-shot learning. (a) The training samples whose first $n$ columns are used for $n$-shot learning. (b)  The weight patterns of the $36$ most active neurons in the first layer of $10$-shot learning. (c) The test error rates of $n$-shot learning with $n=1,2,5,10$ as training progresses. }
\label{fig_fewshot_learning}
\end{figure}

What happens if activation learning encounters even fewer samples, as described by few-shot learning, which addresses the challenge of training a model with a very small number of training samples?  To evaluate the performance of activation learning for few-shot learning, we build a training set, as shown in Fig.  \ref{fig_fewshot_learning}(a), consisting of $10\times 10$ randomly selected images from MNIST. The $n$-shot learning refers to the situation in which each digit is presented with $n$ images from the first $n$ columns of the training set. Fig. \ref{fig_fewshot_learning}(c) illustrates the test accuracy of activation learning without feedback and data augmentation as training advances for $n=1,2,5$ and $10$.
The accuracy of the $1$-shot activation learning test is around $48.4\%$. And with $10$ shots, the accuracy rises to approximately $85.0\%$, surpassing backpropagation with $600$ training samples. It verifies the superiority of activation learning with a small number of training samples. The features learned by the 36 most active neurons in the first layer through $10$-shot learning are depicted in Fig. \ref{fig_fewshot_learning}(b). It is not easy to recognize single digits from the features, consistent with our findings that the local competitive learning rule seeks to discover some non-orthogonal principal components of the input images as opposed to memorizing individual images. This distinguishes activation learning from other bio-inspired learning models  \cite{Krotov2019}.

\subsection{Robustness of Learning}

\begin{figure}[!t]
\centering
\includegraphics[width=5.6in]{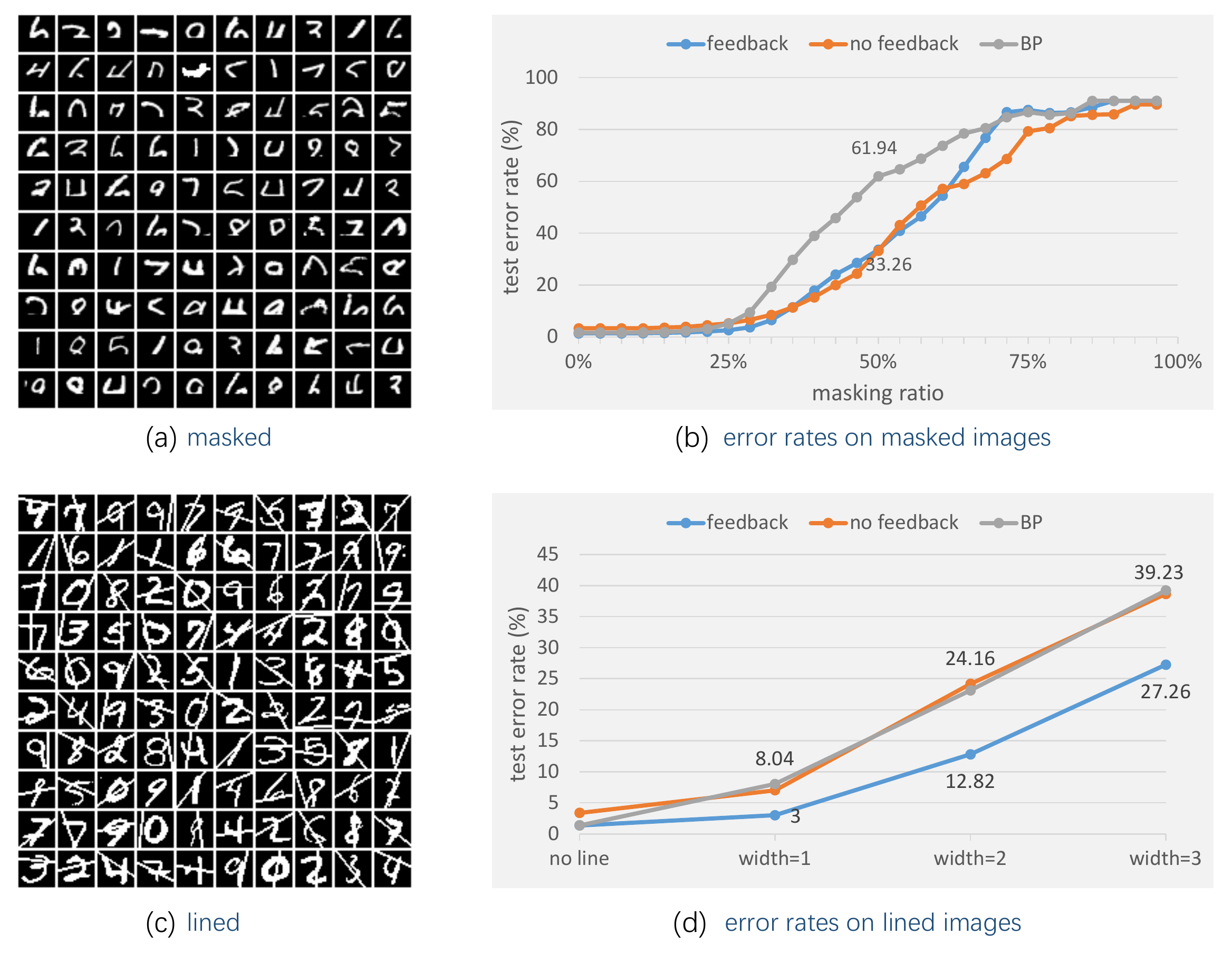}
\caption{(a) Masked images in which only the upper portion is visible. (b) The classification error rates of the trained two-layer network for masked images with various masking ratios.
(c) Images with lines inserted at random. (d) The classification error rates of the trained two-layer network for images with random lines of different widths.}
\label{fig_inference_mask}
\end{figure}

Being robust against external disturbances such as occlusion is essential to human learning, which drives us to study the robustness of activation learning in recognizing incomplete or disrupted images. In the first group of experiments, we mask the bottom  portions ($a \times 28$ pixels) of the test images as black (zero), as seen in  Fig. \ref{fig_inference_mask}(a) for some samples, and recognize the masked images using previously trained neural networks that have not yet learned masked images. Fig. \ref{fig_inference_mask}(b) depicts the test error rates for recognizing masked images with various masking ratios, i.e., the fraction of pixels being masked in each image, $a/28$. It demonstrates that activation learning, with or without feedback, is superior to backpropagation in terms of tolerating portions of the input images being covered. For a masking ratio between 25 and 50 percent, the classification error rate of backpropagation is nearly double that of activation learning.

In the second group of experiments, as depicted in Fig. \ref{fig_inference_mask}(c), we add random lines of a certain width to the test images and study the influence of the line width (in pixels) on the classification error rates. Fig. \ref{fig_inference_mask}(d) reveals that activation learning without feedback achieves performance comparable to backpropagation, while activation learning with feedback achieves the best performance. When the width of the randomly added lines is $1$, activation learning with feedback has a test error of around $3.0\%$, which rises to approximately $12.8\%$ when the width of the lines is $2$. It demonstrates that activation learning with feedback can tolerate a certain level of disturbances in images for recognition, despite the fact that the model was taught nothing about the disturbances.

\subsection{Image Generation and Completion}

In the context of activation learning, image generation is analogous to classification: classification is to infer the categories from the data, whereas image generation is to infer data from nothing or given categories. In the experiments, we generate images for each digit using the same two-layer network previously trained without feedback via activation learning. Applying sampling techniques such as Gibbs sampling based on the estimated probability distribution of the input pattern is a straightforward idea, but they are generally computationally too slow. Here, we generate some images of given digits by maximizing output activations.

\begin{figure}[!t]
\centering
\includegraphics[width=5.0in]{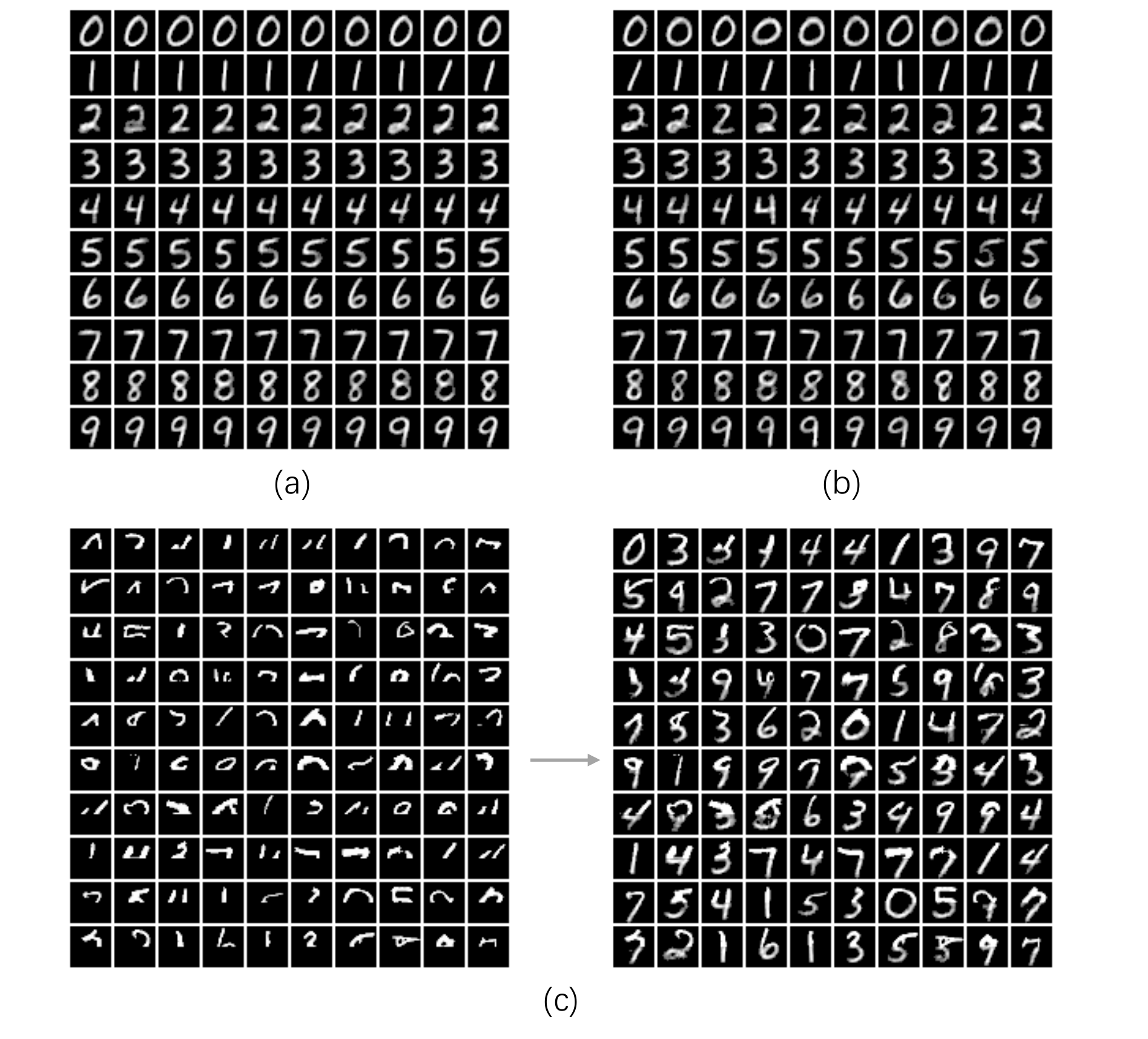}
\caption{(a) Images generated without introducing noise. (b)  Images generated with noise. (c) Masked images and the corresponding completed images.
}
\label{fig_generative_model}
\end{figure}

A $L_1$-norm term is introduced to the objective function of inference for the purpose of sharpening the generated digits:
\begin{equation}
\mathbf{z}^* = \arg\max_{\mathbf{z}} \|\mathbf{y}(\mathbf{a}, \mathbf{z})\|^2 - \beta \|\mathbf{z}\|_1 \label{equ4_1}
\end{equation}
such that $\|\mathbf{z}\|=1$, where $\beta$ is chosen $0.003$. Each image generated, beginning with a uniformly distributed random image, is optimized using the gradient descent method in accordance with (\ref{equ_alternative}). Fig. \ref{fig_generative_model}(a) depicts randomly generated images, which achieve a certain level of quality but not enough diversity. One reason is that they are close to some locally optimal points of the objective function, whose diversity depends solely on the random starting locations and non-convexity of the objective function.

The diversity of the generated images can be increased by infusing randomness into the network that regulates the expression of specific features of the images. One way is to modulate each output of the network by adding some noise so that the resulting output activation is
\begin{equation}\sum_i (1+\delta_i) y_i^2 \end{equation}
where $\delta_i$ is a zero-mean Gaussian noise with standard deviation $0.03$. This noise impacts the contribution of the associated features in the generated images. Fig. \ref{fig_generative_model}(b) shows some newly generated images with better diversity.

Image completion is an additional application of the preceding approach. In the experiments, given masked images with only the top half visible, we attempted to reconstruct the entire image by inference. The task consists of two steps: identifying the digits based on the masked images, and then inferring the missing parts based on the visible parts and the predicted digits. The inference is to solve an unconstrained problem:
\begin{equation}\mathbf{z}_\mathrm{h}^* = \arg\max_{\mathbf{z}_\mathrm{h}} \|\mathbf{y}(\mathbf{a}, \mathbf{z}_\mathrm{v}, \mathbf{z}_\mathrm{h})\|^2,  \end{equation}
where $\mathbf{a}$ is the normalized encoded representation of the label, $\mathbf{z}_\mathrm{v}$ is the given visible part of the image and $\mathbf{z}_\mathrm{h}$ is the hidden or missing part of the image.
$(\mathbf{z}_\mathrm{v}, \mathbf{z}_\mathrm{h})$ is jointly normalized before feeding to the network. Fig. \ref{fig_generative_model}(c) shows some randomly selected masked images and the corresponding completed ones. Even though there is a certain chance of misrecognition based just on the upper half of the images, the majority of the images are  properly reconstructed.

\section{Experiments with CIFAR-10}

\label{section_local_connections}

This section studies activation learning on neural networks with local connections, and conducts experiments on the CIFAR-10 dataset, which consists of $60000$ color images in $10$ categories, with each image containing $32\times 32$ pixels.

\subsection{Network with Local Connections}

\begin{figure}[!t]
\centering
\includegraphics[width=5.4in]{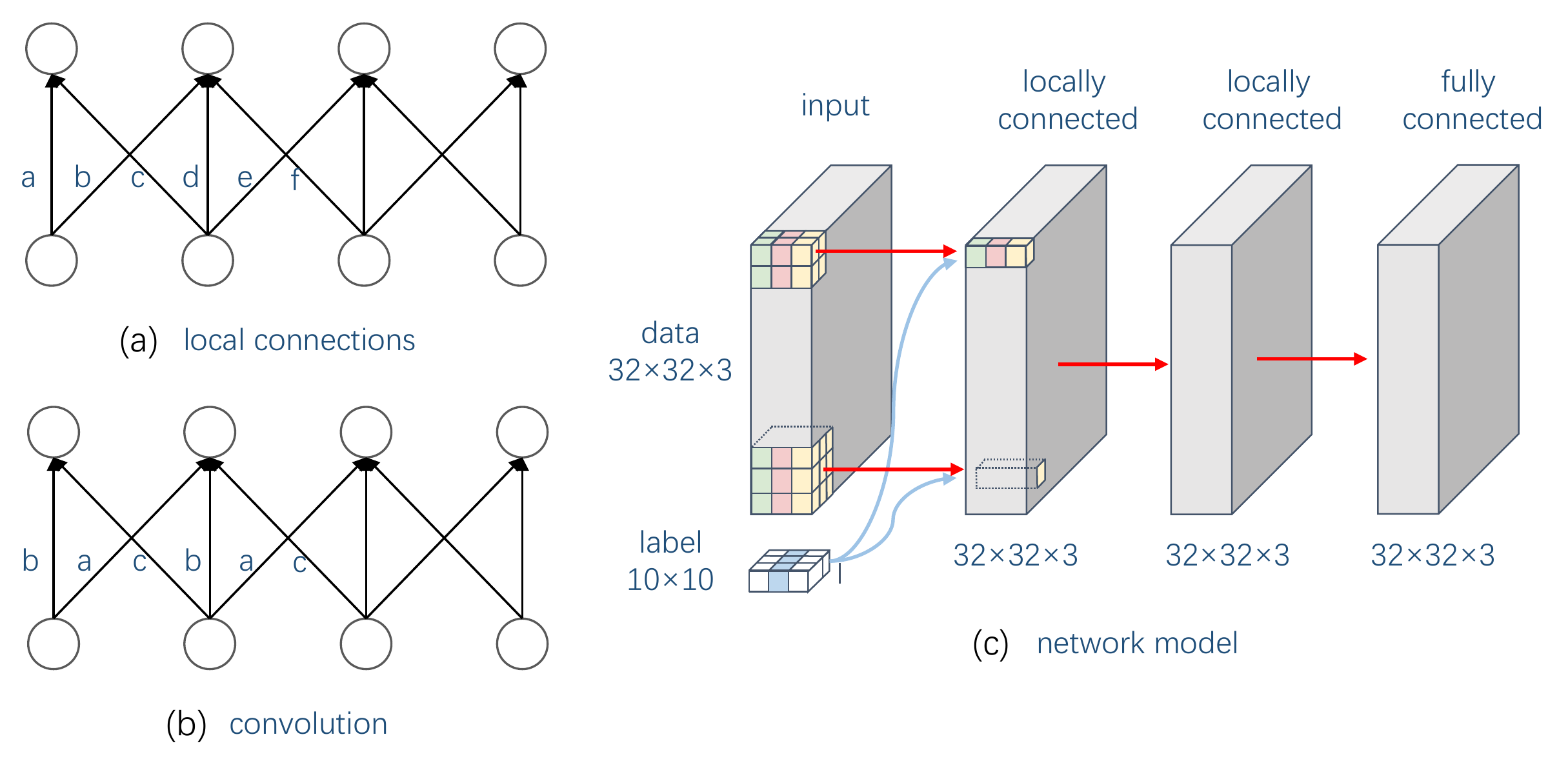}
\caption{(a) A locally connected layer in which each connection has its own weight. (b) A convolutional layer with sharing weight parameters. (c)
A network model with a few locally connected layers and one fully connected layer.}
\label{fig_convolution_filter}
\end{figure}

Convolutional networks \cite{Krizhevsky2012} have been tremendously successful in practical applications of computer vision and time-series processing. Given a convolutional layer, an input pattern is cut into small patches of the same size, which are presented to the same filters for feature detection. The convolution model was originally proposed as a biologically-inspired synthesized model. The purpose of convolutional networks is to reduce the number of model parameters and enhance the recognition of patterns  that have shifted in location.  The parameter sharing mechanism of convolutional layers, however, is not biologically plausible and hence differs from how the human brain works.  For activation learning, we are interested in locally connected layers that have the same structure as convolutional layers but do not share parameters across locations. In  Figs. \ref{fig_convolution_filter}(a) and (b), locally connected layers with each connection having its own weight are compared to convolutional layers. Locally connected layers have more parameters than convolutional layers, but they allow fewer neurons at each location in activation learning and thus require less computation. Moreover, a locally connected layer may capture location-related features, which could be valuable for facial recognition, for instance. Local connections that force neighboring neurons to compete are particularly suited for the discovery of lower-layer features in activation learning.

The network model used in our studies is depicted in Fig. \ref{fig_convolution_filter}(c), which comprises several locally connected layers as the hidden layers and a fully connected layer as the output layer. The input of the network consists of a $32\times 32 \times 3$ normalized image and a $10\times 10$ normalized encoded label (each category is represented by a row). Each network layer consists of $32\times 32 \times 3$ units (the same as an input image) and uses the std\_abs activation function. A neuron unit in a locally connected layer connects to the units in the layer below within a $9 \times 9 $ receptive field. Each neuron in the first layer is not only connected to the local units of the input image but also to all the units of the encoded label. Training a locally connected layer is identical to training a fully connected layer, with the exception that some weights are fixed to zero.
When learning with feedback, the same updating rule is used as for the MNIST experiments, and the unlearning factor is set to $\lambda = 0.7$ ($\lambda = 0$ when the output activation is less than a given threshold for faster starting).

\subsection{Classification Experiments}

\begin{figure}[!t]
\centering
\includegraphics[width=6.0in]{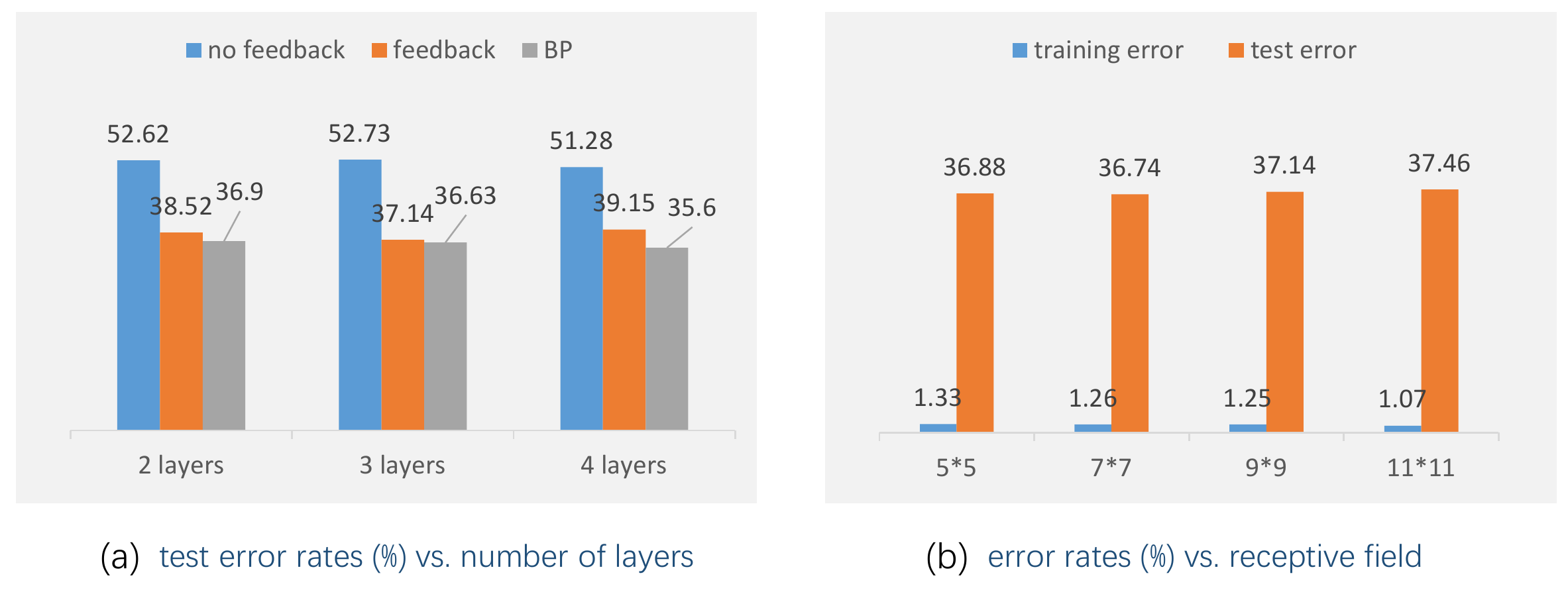}
\caption{(a) The test error rates of a network trained on CIFAR-10 by activation learning or backpropagation. (b) The error rates of a $3$-layer network trained by activation learning with feedback for different receptive fields.}
\label{fig_cifar10_results}
\end{figure}

Fig. \ref{fig_cifar10_results}(a) compares the test error rates of activation learning using a network with two to four layers to those of backpropagation.
The network used for backpropagation has an additional layer with $10$ units on top of the network used for activation learning, and each hidden layer utilizes ReLU activation. Since activation learning without feedback overfits, it uses $5,000$ training samples for validation, which is unnecessary for activation learning with feedback and backpropagation in the experiments. Without feedback, activation learning achieves about a $52\%$ test error rate. And with feedback, this test error rate is reduced to about $37.14\%$ on a three-layer network. It approaches the error rate of backpropagation on a network of three hidden layers very closely. Nevertheless, the error rate does not continue to decrease as network depth increases, which requires additional investigation. A previous benchmark of biologically plausible networks without convolutional layers by Krotov and Hopfield reports a $49.25\%$ error rate \cite{Krotov2019} on CIFAR-10, which used a two-layer network of $2000$ hidden units, but the top layer was still trained by backpropagation. Using a network of the same size as in our experiments, Hinton recently presented the Forward-Forward Algorithm \cite{hinton2022forward}, which is backpropagation-free and yields a $41\%$ test error rate.

Fig. \ref{fig_cifar10_results}(b) examines the impact of receptive field size on the error rates of activation learning when using a three-layer network with feedback. Learning with a fully connected network does not converge with the same hyper-parameters; hence, it is not depicted in the figure. With a $5\times 5$ receptive field, the test error rate can be reduced to approximately $36.88\%$, nearly the same as that of propagation. Network performance can also be enhanced by increasing the network's width; if the receptive field remains $9\times 9$, but the number of units in each layer increases to $32\times 32 \times 10$, the test error rate can reach around $35.6\%$.

\subsection{Learning with Data Augmentation}

In addition to increasing the depth or width of the network, using local connections, and learning with feedback, data augmentation techniques such as randomly cropping can be utilized to enhance the performance of a network. As fully connected layers and locally connected layers are not shift invariant, data augmentation enables the recognition of shifted images without the need for convolutional layers. In the experiments, simple data augmentation techniques, including randomly cropping images to $28\times 28 $ pixels and randomly flipping images horizontally, are applied. To match the cropped images, we use a network of three layers, with each layer  containing  $28\times 28$ locations and using a $5\times 5$ receptive field for local connections.

\begin{figure}[!t]
\centering
\includegraphics[width=6.0in]{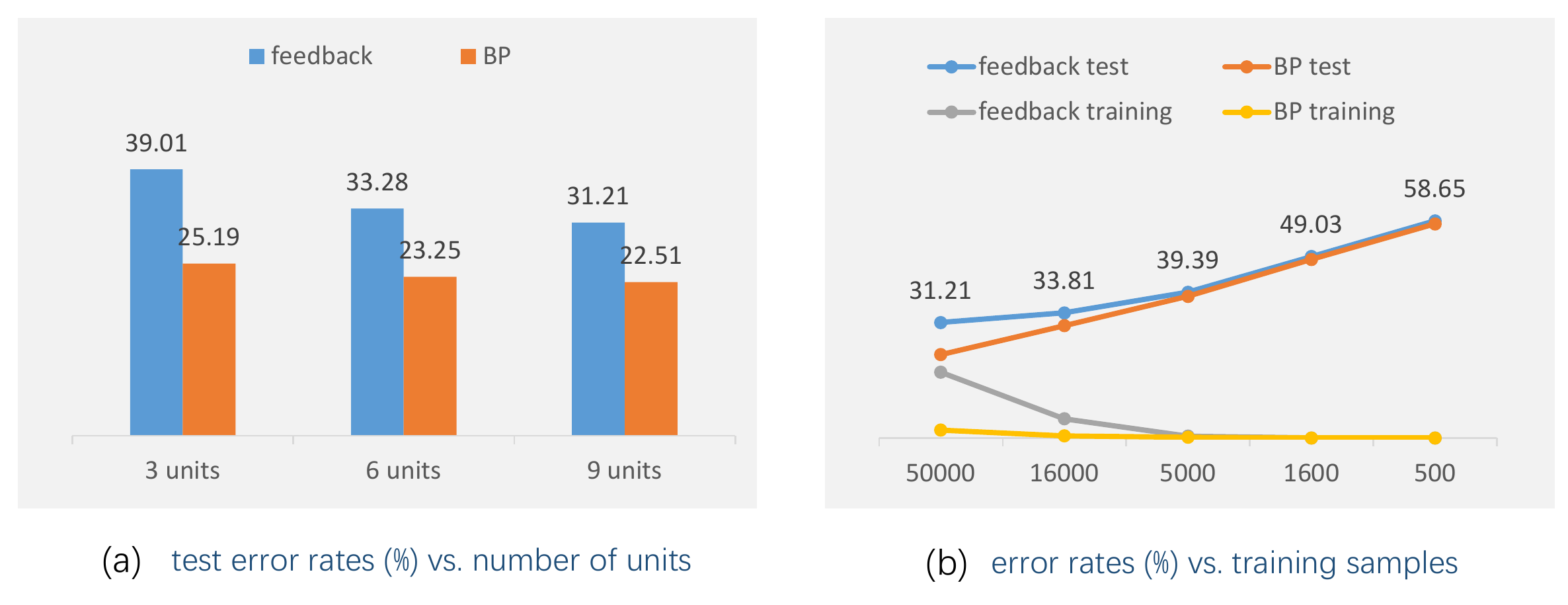}
\caption{(a) The test error rates of a three-layer network trained on CIFAR-10 with data augmentation, where each layer has $28\times 28\times n$ units with $n=3$, $6$, or $9$, respectively. (b) The error rates of a three-layer network trained using a specific number of training samples from CIFAR-10, with $9$ units at each layer location. }
\label{fig_cifar10_augmentation}
\end{figure}

Fig. \ref{fig_cifar10_augmentation}(a) studies how the number of units at each location influences the network's performance when using the simple data augmentation on CIFAR-10. When each location has $3$ units (the same as without data augmentation), data augmentation does not increase network performance, and training is difficult to converge because the network is too narrow for sufficient expressibility. By increasing the number of units per location to $9$, the test error rate of activation learning with feedback using a three-layer network reaches approximately $31.21\%$. There is a performance gap between activation learning and backpropagation when data augmentation is applied. This performance gap may be narrowed by increasing the network's width (the number of neurons in each layer).

Fig. \ref{fig_cifar10_augmentation}(b) depicts the relationship between error rates with simple data augmentation and the number of training samples when each layer has $9$ units at each location. When the number of training samples is decreased, activation learning with feedback does not outperform backpropagation.  This is likely due to the fact that the backgrounds of the CIFAR-10 images are complicated and the images' primary objects are not centered. However, the performance of activation learning with feedback and that of backpropagation do converge when the number of training samples is decreased. It implies that for activation learning to work best, the size of the network needs to match the sample complexity.

%\section{Practical Applications}
%
%Anomaly Detection
%
%A practical method: for example, for anormaly detection, where the output activation smaller than a specified threshold is classified anormly. Our recent work surpass the sota on a variety of real-world databases.
%
%Prediction: not only the value, but also the distribution, useful in many applications.
%
%Small Overfitting: for many applications, do not need validation set!
%
%Could improve $10\%$

\section{Discussions}

\label{section_discussion}

\subsection{Practical Applications}

Activation learning currently cannot compete with backpropagation that employs complex regularizers and convolutional networks for classification and image generation. Scaling activation learning to big neural networks and large datasets requires additional research. Besides of classification and image generation, activation learning can be used in many other tasks. Our subsequent paper by Ding et al. used activation learning for anomaly detection on several real-world datasets, including traffic data \cite{ahmad2017unsupervised} and arterial blood data, and achieved competitive performance compared to state-of-the-art models \cite{2020USAD,2021GND} trained by propagation, but with a simple fully connected  network. The absence of anomalous samples is the greatest obstacle in many applications of anomaly detection, but activation learning can learn from only normal observations without supervision and activate less on anomalous patterns. This makes it a simple and effective method for anomaly detection: an input pattern is classified anomalous if the network's output activation is below a predetermined threshold. It demonstrates that biologically plausible learning models can outperform backpropagation in certain real-world scenarios.

Forecasting is another potential application of activation learning. Most existing forecasting models trained by propagation attempt to predict future time series such as sales, prices, and temperatures; however, very few can predict the complex distribution of future series, or they rely on some simple assumptions such as the independent Gaussian distribution. In scenarios such as financial trading, which is concerned with both the expected return and the overall risk, the ability to accurately predict statistical distributions, as opposed to series of values, is desirable. Activation learning paired with efficient sampling techniques to approximate the distribution of future series could be a powerful forecasting technique.

\subsection{A Unified Framework}

\begin{figure}[!t]
\centering
\includegraphics[width=5.8in]{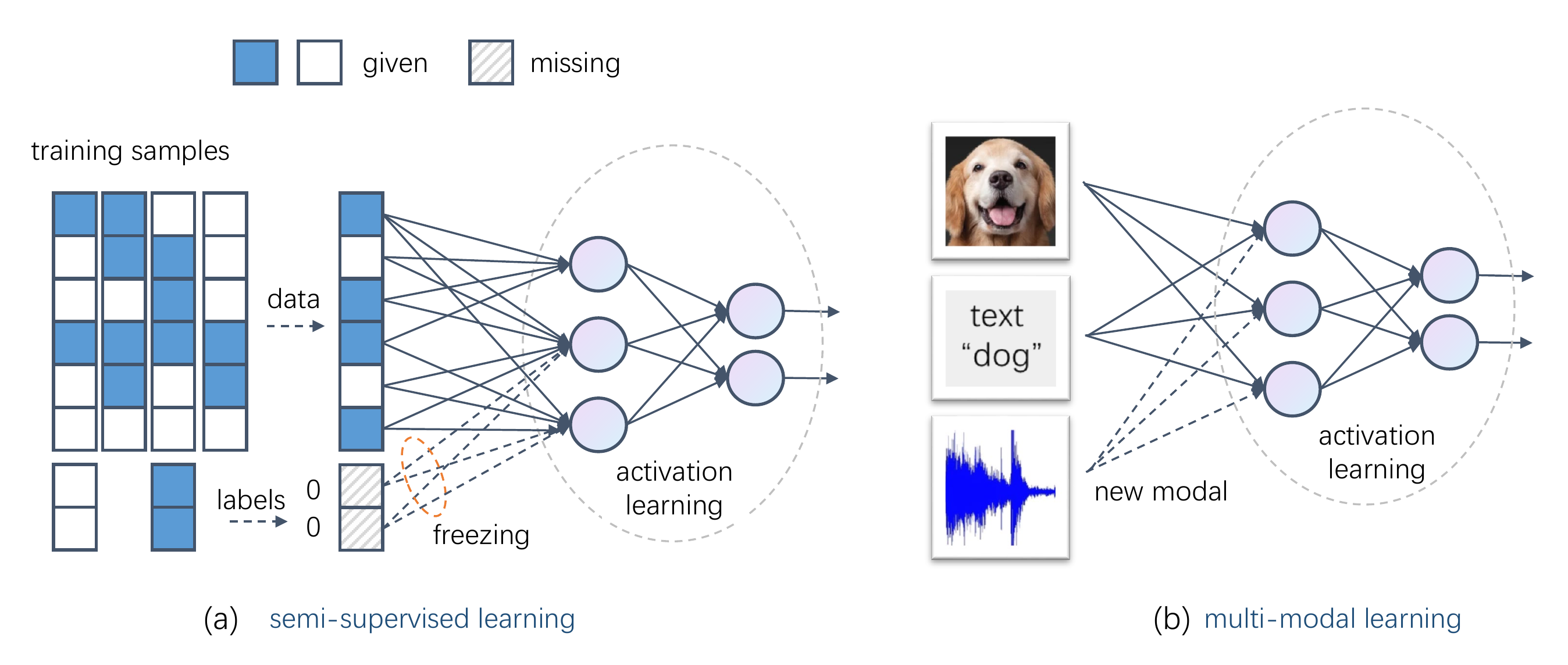}
\caption{(a) Activation learning for semi-supervised learning. (b) Activation learning for multi-modal learning. }
\label{fig_unified_learning}
\end{figure}

The potential for activation learning to unify supervised learning, unsupervised learning, and generative models is appealing. In contrast to a conventional neural network, which takes the data as the input and utilizes the label to calculate the output error, an activation learning neural network uses both the data and label as inputs. However, labeling data is costly in many scenarios, and it is desired to discover some valuable features from unlabeled data in order to improve the performance of some learning tasks, which leads to unsupervised learning, self-supervised learning, and semi-supervised learning. As depicted in Fig. \ref{fig_unified_learning}(a), supervised learning and unsupervised learning may be directly integrated in activation learning. When encountering unlabeled data, it is possible to feed them directly to the learning network while deactivating the label input and some related neurons in order to learn some unsupervised data features. Meanwhile, when meeting data with labels, the supervised label information can help reshape the unsupervised features and establish their association with the label.

It poses a general and unresolved challenge, namely, how to learn from partial units of training samples, i.e., based on any subset of input units at each training step. For example, given a training sample with some missing units, the input pattern can be completed by filling the missing units with zeros (or other values), and training can be conducted by updating some connection weights while freezing those associated with the missing units. Consequently, it is feasible to learn either from the data alone or from the data-label pair, depending on whether the label is available for a particular sample.

This framework of activation learning can also be extended to multi-modal learning when multiple modals are used as inputs; see Fig. \ref{fig_unified_learning}(b). It is desirable for the network to be able to be trained by any subset of the modals at any one time and to establish associations between the modals.  It is considered that such an associative ability, which leads to associative memory, is crucial to human intelligence \cite{matzen2015effects}. For example, if the activation learning network receives both
the images and texts of some objects, it may learn to establish associations between the images and texts.  Similarly, providing additional pairs of sounds and texts may establish associations between sounds and texts. If statistically related items (which often appear together) can induce activation learning to build their associations as a graph, the associations may spread to create relationships between unrelated items. An important challenge is how to add new inputs (such as additional modals) to a well-trained neural network without decreasing its performance on existing tasks. Existing models of deep learning struggle to handle this fundamental obstacle to general-purpose intelligence.  But activation learning seems to be a promising framework to handle it.

\subsection{A General-Task Paradigm}

\begin{figure}[!t]
\centering
\includegraphics[width=5.6in]{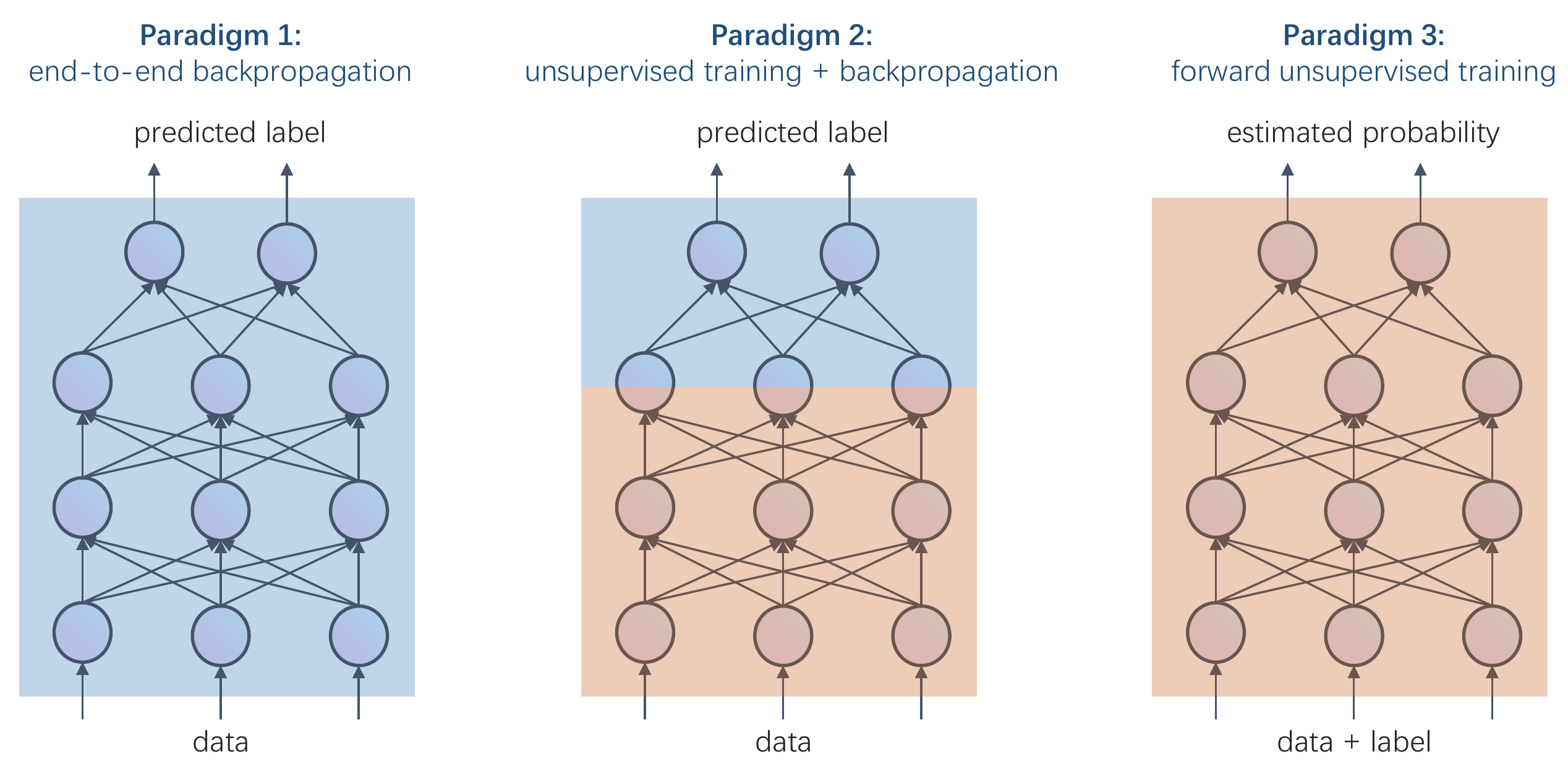}
\caption{Three learning paradigms: (a) End-to-end training by backpropagation. (b) Bottom-up unsupervised training of lower layers and then training top layers with supervision.  (c) Bottom-up unsupervised training of all the layers to estimate the joint probability of the input.}
\label{fig_learning_paradigms}
\end{figure}

Three general learning paradigms are shown in Fig. \ref{fig_learning_paradigms}. The first paradigm is end-to-end backpropagation training of the network.  The second paradigm is to train the lower layers without supervision and then supervise the training of the top layer or layers. Most biologically plausible learning models and unsupervised pre-training models belong to the second paradigm. Nonetheless, the second paradigm faces the issue that the learned unsupervised features may not be a perfect match for the supervised learning task.  Consequently, it is usually required to fine-tune the entire network in order to reshape the unsupervised features and improve network performance. The third paradigm, activation learning, uses both the data and the label as input, trains all layers in an unsupervised and forward manner, and outputs a score that reflects the relative likelihood of the input pattern. The third paradigm, which employs multiple layers, enables the discovery of complex correlations between data and label.  Due to the fact that the underlying learning rule of activation learning attempts to explore the inherent correlations of the input patterns regardless of specific tasks, the network models can be used for general purposes.

The first and second paradigms are directional for data and labels, with either a discriminative model using the data as input and the label as output supervision or a generative model using the label as input to infer the data. If one already possesses a discriminative model and wants to generate data patterns from given labels, a new generative model needs to be created.  In addition, diverse data sources necessitated a large number of models to meet a variety of inference requirements (from any subset of sources to the others),  which is not feasible for the brain. Activation learning accepts all data as inputs (including labels) and learns their correlations with a single model, enabling bidirectional inference between any input parts. If asked what information a model should learn from data samples without knowing future tasks, our answer would be the most precise possible statistical distribution of the data samples.

%\subsection{Inspirations about Brain}
%
%It is unclear that activation learning is only a brain-inspired mathematical model, or it may bring us some inspirations about understanding the brain.
%
%(1) Whether neuron output has a inhabiting signal to all its input units?
%
%(2) Whether there are some mechanism in brain that is similar to the $abs$ activation?
%
%(3) Whether brain can implement unlearning?
%
%(4) Whether brain imaging activates more on familiar instances than unfamiliar instances?

\subsection{Relationship to Forward-Forward Algorithm}

The Forward-Forward algorithm\footnote{Several months after the original release of activation learning on arXiv (\href{https://arxiv.org/abs/2209.13400}{https://arxiv.org/abs/2209.13400}), Hinton posted his unfinished paper on the Forward-Forward Algorithm on his website (\href{https://www.cs.toronto.edu/\~hinton/FFA13.pdf}{https://www.cs.toronto.edu/\%7Ehinton/FFA13.pdf}). Two researchers independently arrived at some similar key ideas. } very recently invented by Hinton \cite{hinton2022forward} presents some interesting and similar key ideas as activation learning. Starting from a local bio-plausible learning rule, activation learning derived that the sum of the squared output can be used as a good estimate of the input probability, whereas the Forward-Forward algorithm (FF) used the sum of the squared output as the goodness function for its simplicity and then derived a training method.  Both of the works used labels as input,  trained networks with both positive and negative samples, and employed the same locally connected layers without parameter sharing. FF accumulates the activations of all but the first hidden layer as goodness while normalizing the input of each layer. But activation learning directly measures likelihood using the top layer's output activation, yielding a score $h_1(\mathbf{x}) h_2(\mathbf{x})  ... $ with $h_i(\mathbf{x})$ is the activation filter of the $i$th layer. As a comparison, accumulating the activations of all the layers while normalizing the input of each layer yields $h_1(\mathbf{x}) + h_2(\mathbf{x}) + ... $, which is the first-order approximation of $h_1(\mathbf{x}) h_2(\mathbf{x})  ... $ as $h_i(\mathbf{x})$ is close to $1$ in activation learning. These coincidental discoveries encourage us to believe that this direction, i.e., estimating input distribution (likelihood or goodness) with unsupervised forward learning, is worthy of in-depth investigation.

There are a number of distinctions between activation learning and FF. (1) They have different underlying unsupervised updating rules to train each layer. Activation learning uses  the local competitive learning rule that causes the output activation to be upper-bounded by the input strength, whereas the output activation in FF is not bounded.  (2) Activation learning trains all layers concurrently with feedback (i.e., when one layer can distinguish between positive and negative samples, the other layers are not trained further on these samples), whereas the FF algorithm trains each layer independently. (3) FF normalizes the output of each layer before feeding it to the layer above. This differs from activation learning, in which higher layers are less active than lower layers. (4) Activation learning can learn without negative samples, making it easier to use in applications like anomaly detection, where it achieves cutting-edge performance on several real-world datasets.

\section{Conclusion}
\label{section_conclusion}

We believe that the brain is a complex system governed by a simple rule and dominated by unsupervised learning. This inspires the development of activation learning, a new learning paradigm for bottom-up unsupervised learning based on the local competitive learning rule, and whose output activation estimates the likelihood of input patterns. On the MNIST and CIFAR-10, activation learning with feedback of accuracy information that generates negative samples for the network to unlearn achieves comparable classification performance to backpropagation using a simple network. Activation learning exhibits a number of benefits that make it a potential complement to backpropagation. (1) The model of activation learning is generic, and the same network can be used for a variety of learning tasks, such as classification, prediction, detection, generation, and completion. (2)  Activation learning has demonstrated its superiority in classification in terms of learning from fewer samples, robustness against external disturbances, and resistance to adversarial attacks. (3) Activation learning excels in unsupervised learning for estimating input distribution, allowing it to achieve state-of-the-art performance for anomaly detection across several real-world datasets. (4) Activation learning has the potential to unify supervised learning, unsupervised learning, and multi-modal learning and may therefore contribute to the development of artificial general intelligence. (5) Activation learning's local learning rule makes it easier to implement on neuromorphic computing hardware for on-chip training and may overcome the von Neumann architecture's efficiency and scaling bottlenecks. Despite a number of challenges that need to be handled, such as learning from partial units, the absence of fast inference algorithms, and the training of deep neural networks, activation learning emerges as a promising learning paradigm deserving of extensive research.

\section*{Acknowledgement}

We would like to acknowledge Weiping Huang and Fei Liu for their discussions and feedback, and acknowledge Fengqian Ding for his contributions on anomaly detection based on activation learning. The code for activation learning has been open-sourced \footnote{\href{https://github.com/DPSpace/activation\_learning}{https://github.com/DPSpace/activation\_learning}}, making it easier to build upon this work.

\appendices

\section{Local Competitive Learning}
\label{Appendix_rule}

We analyze the converged properties of the local competitive learning rule (\ref{equ2}), starting with a study of its dynamics.  The net input of the neurons in a layer is a linear transformation of the input pattern, denoted by \begin{equation}\mathbf{y}=\mathbf{w}^T\mathbf{x}\end{equation}
with $\mathbf{y}=[y_1, y_2, ...]$ the net input vector, $\mathbf{x}=[x_1, x_2, ...]$ the input vector, and $\mathbf{w}$ the connection weight matrix.

The rewritten version of the local competitive learning rule is
\begin{equation}
\Delta w = \eta  (\mathbf{x} \mathbf{x}^T \mathbf{w} - \mathbf{w} (\mathbf{w}^T \mathbf{x} \mathbf{x}^T \mathbf{w})).
\end{equation}
With with a very small learning rate $\eta$, the dynamics of $\mathbf{w}$ can be approximated by the differential equation
\begin{equation}
\frac{d\mathbf{w}}{dt} = C \mathbf{w} - \mathbf{w} (\mathbf{w}^TC \mathbf{w}),\label{equ3}
\end{equation}
where $\mathbf{w}$ stands for a time-dependent matrix $\mathbf{w}(t)$ and $C=E\{\mathbf{x} \mathbf{x}^T\}$ is the covariance matrix of the training samples.
The covariance matrix $C$ is  positive semidefinite, i.e., all its eigenvalues are positive.

The connection weights $\mathbf{w}$  tend to converge to asymptotically stable solutions of (1) given by \begin{equation}
 C \mathbf{w} - \mathbf{w} \mathbf{w}^T C \mathbf{w} = 0. \label{equ4}
\end{equation} after a sufficient number of training steps.
When the number of receiving neurons $m\geq 2$, the number of stable solutions is infinite. Which stable point the local competitive learning rule converges to depends on the initial value of $\mathbf{w}$, the training input patterns, and the learning rate. From condition (\ref{equ4}) of the stable solutions,  we can derive a number of results that aid in the comprehension of the proposed learning rule.

\subsection{Reconstruction Error}

\textbf{Property 1}: The local competitive learning rule performs to approximately minimize the mean reconstruction error
\begin{equation}E\|\mathbf{x} - \mathbf{w} \mathbf{w}^T \mathbf{x}\|^2,\label{equ2_1}\end{equation}
where $\mathbf{w} \mathbf{w}^T \mathbf{x} = \mathbf{wy}$ is considered a reconstruction of $\mathbf{x}$. It follows that the gradient of this reconstruction error with respective to $\mathbf{w}$ equals to $0$ when $\mathbf{w}$ is a stable solution satisfying (\ref{equ4}). This implies that the learning rule attempts to decompose the input pattern $\mathbf{x}$ into $m$ components with the minimum mean reconstruction error.

\begin{proof}
The mean reconstruction error
$$\|\mathbf{x} - \mathbf{w} \mathbf{w}^T \mathbf{x}\|^2 = \sum_{i} (x_i - \sum_{j,k}w_{ij}w_{kj}x_k)^2$$
where $i, k$ enumerates the input units and $j$ enumerates the output units.

If we calculate the partial derivative on the reconstruction error with regard to $w_{uv}$, it has
\begin{eqnarray*}
  \frac{\partial \|\mathbf{x} - \mathbf{w} \mathbf{w}^T \mathbf{x}\|^2}{\partial w_{uv}} &=& - 2 \sum_i (x_i - \sum_{j,k}w_{ij}w_{kj}x_k) ((\sum_{k} w_{kv}x_k) I_{i=u} + w_{iv} x_u)\\
   &=& - 2 (\sum_k x_u x_k w_{kv} - \sum_{j,k, l}w_{uj}w_{kj}x_k x_l w_{lv}) \\
   & & - 2 (\sum_i x_u x_i w_{iv} - \sum_{i, j, k}x_u x_k w_{kj} w_{ij}  w_{iv}).
\end{eqnarray*}

It results in
\begin{equation}\frac{dE\|\mathbf{x} - \mathbf{w} \mathbf{w}^T \mathbf{x}\|^2}{d\mathbf{w}}  =
 -2[C \mathbf{w} - \mathbf{w} \mathbf{w}^TC \mathbf{w}] - 2[C\mathbf{w} - C\mathbf{w} \mathbf{w}^T \mathbf{w}]\label{equ_appendix_1}
\end{equation}
with the covariance matrix $C = E\{\mathbf{x}\mathbf{x}^T\}$. We further show that if $C \mathbf{w} - \mathbf{w} \mathbf{w}^TC \mathbf{w}=0$, then $C\mathbf{w} - C\mathbf{w} \mathbf{w}^T \mathbf{w}=0$, and hence the gradient (\ref{equ_appendix_1}) becomes $0$. The proof is as follows. If $C \mathbf{w} - \mathbf{w} \mathbf{w}^TC \mathbf{w}=0$, then $CC \mathbf{w}-C\mathbf{w} \mathbf{w}^TC \mathbf{w}=0$, and
\begin{eqnarray*}
  CC \mathbf{w}-C\mathbf{w} \mathbf{w}^TC \mathbf{w} &=& C(\mathbf{w} \mathbf{w}^TC \mathbf{w}) - C\mathbf{w} \mathbf{w}^T(\mathbf{w} \mathbf{w}^TC \mathbf{w})  \\
  &= & (C\mathbf{w} - C\mathbf{w} \mathbf{w}^T \mathbf{w}) \mathbf{w}^TC \mathbf{w}.
\end{eqnarray*}
So either $C\mathbf{w} - C\mathbf{w} \mathbf{w}^T \mathbf{w}= 0$ or $\mathbf{w}^TC \mathbf{w}=0$. If $\mathbf{w}^TC \mathbf{w}=0$, according to the condition that $C \mathbf{w} - \mathbf{w} \mathbf{w}^TC \mathbf{w}=0$, $C\mathbf{w}=\mathbf{w} (\mathbf{w}^TC \mathbf{w})= 0$. This also leads to $C\mathbf{w} - C\mathbf{w} \mathbf{w}^T \mathbf{w}= 0$.

Finally, we reach the conclusion that if $C \mathbf{w} - \mathbf{w} \mathbf{w}^TC \mathbf{w}=0$, then the gradient (\ref{equ_appendix_1}) becomes $0$ and such $\mathbf{w}$ is a local minimum point of the mean reconstruction error $E\|\mathbf{x} - \mathbf{w} \mathbf{w}^T \mathbf{x}\|^2$.
\end{proof}

\subsection{Non-orthogonal PCA}
\label{appendix_pca}

\textbf{Property 2}: For any stable solution $\mathbf{w}$ with $m\geq 1$ receiving neurons, the reconstruction error (\ref{equ2_1}) is equal to that of the principal component analysis (PCA) with $m$ components. This implies that the local competitive learning rule functions similarly to PCA, which ensures that as much information as possible is passed to the next layer. Existing research has shown that PCA can be used in neural networks to extract unsupervised features \cite{Chan2015}. However, unlike PCA, orthogonality is unnecessary for the learning rule's stable solutions. In fact, it is undesirable to use the original PCA to extract features from a layer since the neural network may become vulnerable to the failure of a few neurons corresponding to the largest eigenvalues, and it is difficult to achieve perfect orthonomalization using a local learning rule.

\begin{proof}
We show that the minimum mean square error $E\|\mathbf{x} - \mathbf{w} \mathbf{w}^T \mathbf{x}\|^2$ equals to the reconstruction error of the principal component analysis (PCA).
The proof begins with the observation that $\|\mathbf{x} - \mathbf{w} \mathbf{w}^T \mathbf{x}\|\geq \|\mathbf{x} - \mathbf{w} \mathbf{a}_\mathbf{x}\|$ if $\mathbf{a}_\mathbf{x}$ is independently optimized for each $\mathbf{x}$. The reconstruction error of the PCA is equal to the minimized value of $E\|\mathbf{x} - \mathbf{w} \mathbf{a}_{\mathbf{x}}\|^2$. This shows that the mean square error $E\|\mathbf{x} - \mathbf{w} \mathbf{w}^T \mathbf{x}\|^2$ cannot be smaller than the PCA's reconstruction error.

According to the theory of the PCA, $E\|\mathbf{x} - \mathbf{w} \mathbf{a}_{\mathbf{x}}\|^2$ is minimized if $\mathbf{w}= \mathbf{w}^*=[\mathbf{v}_1, \mathbf{v}_2, ..., \mathbf{v}_m]$ with $\mathbf{v}_j$ the eigenvector corresponding to the $j$th largest eigenvalue of the covariance matrix $C=\{\mathbf{x}\mathbf{x}^T\}$ and $\mathbf{a}_{\mathbf{x}} = {\mathbf{w}^*}^T \mathbf{x}$.  This leads to the conclusion that the reconstruction error of the PCA equals to $E\|\mathbf{x} - \mathbf{w}^* {\mathbf{w}^*}^T \mathbf{x}\|^2$.
Therefore, the minimum mean square error $E\|\mathbf{x} - \mathbf{w} \mathbf{w}^T \mathbf{x}\|^2$ does not exceed the PCA's reconstruction error.

Consequently, the minimum mean square error $E\|\mathbf{x} - \mathbf{w} \mathbf{w}^T \mathbf{x}\|^2$ with $m\geq 1$ neurons is equal to the reconstruction error of PCA with $m$ components.

\end{proof}

\subsection{Stable Solutions}

\textbf{Property 3}: There are infinitely many stable solutions of (\ref{equ3}) that form a surface when the number of neurons $m>1$. Let $\mathbf{v}_1, \mathbf{v}_2, ..., \mathbf{v}_m$ be the normalized eigenvectors corresponding to the $m$ largest eigenvalues of the covariance matrix $C = E\{\mathbf{x}\mathbf{x}^T\}$, then the weight matrix $\mathbf{w}$ is a stable solution if and only if for any $1\leq j\leq m$, the $j$th column (the weights of neuron $j$) of $\mathbf{w}$ can be written as
\begin{equation}\mathbf{w}_j = u_{j1} \mathbf{v}_1 + u_{j2} \mathbf{v}_2 + ... + u_{jm} \mathbf{v}_m\end{equation}
with some $\{u_{jk}\}$ such that $u_{1k}^2 + u_{2k}^2 + ... + u_{mk}^2 = 1$ for all $1\leq k \leq m$ (the factors associated to each eigenvector $\mathbf{v}_k$ is normalized). As a consequence, $\|\mathbf{w}\|= \sum_{ij} w_{ij}^2$ tends to converge to $m$ even though there is no explicit constraint or normalization in the learning rule. This result leads to the conclusion that changing any column $\mathbf{w}_j$ to $-\mathbf{w}_j$ does not affect the stability of a solution $\mathbf{w}$.

\begin{proof}

For any stable solution $\mathbf{w}$ of (\ref{equ4}), each column $\mathbf{w}_j$ can be written as a linear combination of $\mathbf{v}_1, \mathbf{v}_2, ..., \mathbf{v}_m$ with $\mathbf{v}_k$ the eigenvector corresponding to the $k$th largest eigenvalue of the covariance matrix $C= E\{\mathbf{x}\mathbf{x}^T\}$. The reason for this is that any stable solution $\mathbf{w}$ must minimize the reconstruction error $E\|\mathbf{x} - \mathbf{w} \mathbf{w}^T \mathbf{x}\|^2$. It must also minimize $E\|\mathbf{x} - \mathbf{w} \mathbf{a}_\mathbf{x}\|^2$ when $\mathbf{a}_\mathbf{x}$ is optimized independently for each $\mathbf{x}$, which equals to the mean square distance of the samples to the vector space spanned by the columns of $\mathbf{w}$, based on the discussion in Appendix \ref{appendix_pca}. According to the PCA theory, the optimal vector space to minimize this mean square distance must be the one spanned by $\mathbf{v}_1, \mathbf{v}_2, ..., \mathbf{v}_m$ with $\mathbf{v}_j$ the eigenvector corresponding to the $j$th largest eigenvalue of the covariance matrix $C=\{\mathbf{x}\mathbf{x}^T\}$. As a result, each column of a stable $\mathbf{w}$ must be a linear combination of $\mathbf{v}_1, \mathbf{v}_2, ..., \mathbf{v}_m$.

When the number of neurons $m=2$, let a solution $\mathbf{w}=[\mathbf{w}_1, \mathbf{w}_2]$ be
$$\mathbf{w}_1 = u_{11} \mathbf{v}_1 + u_{12} \mathbf{v}_2,$$
$$\mathbf{w}_2 = u_{21} \mathbf{v}_1 + u_{22} \mathbf{v}_2.$$
We show that the solution $\mathbf{w}$ is stable if and only if $u_{11}^2 + u_{21}^2 = 1$ and $u_{12}^2 + u_{22}^2 = 1$. The argument can be naturally extended to a general $m$.

If $\mathbf{w}$ is stable, then $C\mathbf{w} = \mathbf{w}\mathbf{w}^T C\mathbf{w}$, where
\begin{equation}C\mathbf{w} = [\lambda_1 u_{11} \mathbf{v}_1 + \lambda_2 u_{12} \mathbf{v}_2, \lambda_1 u_{21} \mathbf{v}_1 + \lambda_2 u_{22} \mathbf{v}_2 ],\end{equation}
\begin{eqnarray}
\mathbf{w}\mathbf{w}^T C\mathbf{w} &=& ((u_{11}^2 + u_{21}^2) \mathbf{v}_1\mathbf{v}_1^T + (u_{12}^2 + u_{22}^2) \mathbf{v}_2\mathbf{v}_2^T )C\mathbf{w} \nonumber \\
&=& [(u_{11}^2 + u_{21}^2)\lambda_1 u_{11} \mathbf{v}_1 + (u_{12}^2 + u_{22}^2) \lambda_2 u_{12} \mathbf{v}_2, \lambda_1 u_{21} \mathbf{v}_1 + \lambda_2 u_{22} \mathbf{v}_2 ].
\end{eqnarray}
From which, we can get that $u_{11}^2 + u_{21}^2 = 1$ and $u_{12}^2 + u_{22}^2 = 1$. Inversely, if $u_{11}^2 + u_{21}^2 = 1$ and $u_{12}^2 + u_{22}^2 = 1$, we can also deduce that
$C\mathbf{w} = \mathbf{w}\mathbf{w}^T C\mathbf{w}$. In this case, $\mathbf{w}$ is stable, leading to the result stated above. This conclusion also indicates that when $m > 1$, there are an infinite number of stable solutions.

Based on this result, we can also get that for a stable solution $\mathbf{w}$ with $m=2$,
$$\|\mathbf{w}\|^2 = \sum_{ij} w_{ij}^2 =\mathbf{w}_1^T \mathbf{w}_1 + \mathbf{w}_2^T \mathbf{w}_2 = u_{11}^2 + u_{21}^2 + u_{12}^2 + u_{22}^2 = 2.$$
Extending the proof to a general $m$ yields
\begin{equation}\|\mathbf{w}\|^2 = m.\end{equation}
This completes the proof.
\end{proof}

\subsection{Activation Bound}

\textbf{Property 4}: A property of the local competitive learning rule is that the output activation  $\|\mathbf{y}\|^2=\sum_j y_j^2=\|\mathbf{w}^T\mathbf{x}\|^2$ tends to be upper bounded by the input strength $\|\mathbf{x}\|^2$. Let $\mathbf{w}_j$ be the $j$th column of $\mathbf{w}$, in each modification step of the learning rule
\begin{equation}\Delta (y_j^2) \leq y_j^2 ( \|\mathbf{x}\|^2 - \|\mathbf{y}\|^2) + O(\eta^2).\end{equation}
As long as $\|\mathbf{y}\|^2$ is greater than $\|\mathbf{x}\|^2$, $\Delta (y_j^2)$ is smaller than $0$, driving  $\|\mathbf{y}\|^2$ towards smaller than $\|\mathbf{x}\|^2$.

\begin{proof}

We show that with the local competitive learning rule, $\|\mathbf{y}\|^2=\|\mathbf{w}^T\mathbf{x}\|^2$ tends to be upper bounded by $\|\mathbf{x}\|^2$. Let $\mathbf{w}_j$ be the $j$th column of $\mathbf{w}$, from (\ref{equ3}), it derives
\begin{align*}
\Delta (y_j^2) &= \Delta (\mathbf{w}_j^T \mathbf{x})^2 \\
&= 2\eta \mathbf{w}_j^T \mathbf{x} \mathbf{x}^T (\mathbf{x}\mathbf{x}^T\mathbf{w}_j -\mathbf{w}(\mathbf{w}^T\mathbf{x}\mathbf{x}^T\mathbf{w}_j)) + O(\eta^2)\\
&= 2\eta[\mathbf{w}_j^T \mathbf{x} \mathbf{x}^T\mathbf{x}\mathbf{x}^T\mathbf{w}_j - \sum_k (\mathbf{w}_k^T\mathbf{x}\mathbf{x}^T\mathbf{w}_j)^2] + O(\eta^2)\\
&\leq 2 \eta [(\mathbf{w}_j^T\mathbf{x})^2\|\mathbf{x}\|^2 - (\mathbf{w}_j^T\mathbf{x})^2(\sum_k (\mathbf{w}_k^T\mathbf{x})^2)] + O(\eta^2) \\
&= 2\eta y_j^2 ( \|\mathbf{x}\|^2 - \|\mathbf{y}\|^2) + O(\eta^2).
\end{align*}
As long as $\|\mathbf{y}\|^2$ is larger than $\|\mathbf{x}\|^2$, $\Delta (y_j^2)$ is smaller than $0$ if the learning rate $\eta$ is sufficiently small, driving $y_j^2$ to be smaller. As a result, $\|\mathbf{y}\|^2$ tends to be upper bounded by $\|\mathbf{x}\|^2$.
\end{proof}

\subsection{Activation as Typicality}

\textbf{Property 5}: Given a layer trained by the local competitive learning rule, we see that if the input strength $\|\mathbf{x}\|^2$ is fixed, a more probable input causes larger output activation $\|\mathbf{y}\|^2$. This implies that the output activation can be used to estimate the typicality of the input pattern.

To understand the idea behind this,  we assume that $m$ is smaller than the input dimension and denote $V$ as the vector space spanned by the eigenvectors $\mathbf{v}_1, \mathbf{v}_2, ..., \mathbf{v}_m$ corresponding to the largest eigenvalues of the covariance matrix $C=E\{\mathbf{x} \mathbf{x}^T\}$. We refer to $V$ as the typical space. Let $d(\mathbf{x}, \mathbf{V})$ be the distance of $\mathbf{x}$ to the typical space $V$, then
\begin{equation}d(\mathbf{x}, V) = \sqrt{\|\mathbf{x}\|^2 - \sum_i \|\mathbf{x}^T \mathbf{v}_i\|^2}.\end{equation}
We find that, $\|\mathbf{y}\|^2$ tends to be upper bounded by \begin{equation}\|\mathbf{x}\|^2 - d^2(\mathbf{x}, V).\end{equation}
For the special case that $\mathbf{w}=[\mathbf{v}_1, ..., \mathbf{v}_m]$, $\|\mathbf{y}\|^2$ converges to $\|\mathbf{x}\|^2 - d^2(\mathbf{x}, V)$.
Therefore, the output activation $\|\mathbf{y}\|^2$ approximately measures how close the input $\mathbf{x}$ is to the typical space $V$, i.e., the likelihood
of $\mathbf{x}$. Despite the discussion focuses on the case where $m$ is less than the input dimension, the conclusion is is general, which motivates the emergence of activation learning.

\begin{proof}

We show that the output activation $\|\mathbf{y}\|^2$ tends to be upper bounded by $\|\mathbf{x}\|^2 - d^2(\mathbf{x}, V)$ when the output dimension $m$ is smaller than the input dimension, where $d(\mathbf{x}, V)$ is the distance from $\mathbf{x}$ to the linear space $V$ spanned by the eigenvectors $\mathbf{v}_1, \mathbf{v}_2,..., \mathbf{v}_m$ corresponding to the $m$ largest eigenvalues of the covariance matrix $C=\{\mathbf{x}\mathbf{x}^T\}$.

The idea is to construct a new input $\mathbf{x}'$, which is the projection of $\mathbf{x}$ onto the linear space $V$. It is
$$\mathbf{x}'= \sum_{k=1}^m (\mathbf{x}^T \mathbf{v}_k) \mathbf{v}_k.$$
It is easy to verify that when $\mathbf{x}'$ is substituted for $\mathbf{x}$, the net input remains $\mathbf{y}$. According to our result regarding the bound of the output activation, $\|\mathbf{y}\|^2$ tends to be bounded by
$$\|\mathbf{x}'\|^2 = \sum_{k=1}^m (\mathbf{x}^T \mathbf{v}_k)^2 = \|\mathbf{x}\|^2 - d^2(\mathbf{x}, V).$$
This completes the proof.
\end{proof}




\section{Unsupervised Feature Learning}

\label{section_pretraining}

In this section, we conduct experiments on the MNIST dataset to study the features learned by the local competitive learning rule without supervision. We also look at the pre-training models that were built using the learning rule to improve classification performance.

\subsection{Unsupervised Features}

\begin{figure}[!t]
\centering
\includegraphics[width=6.0in]{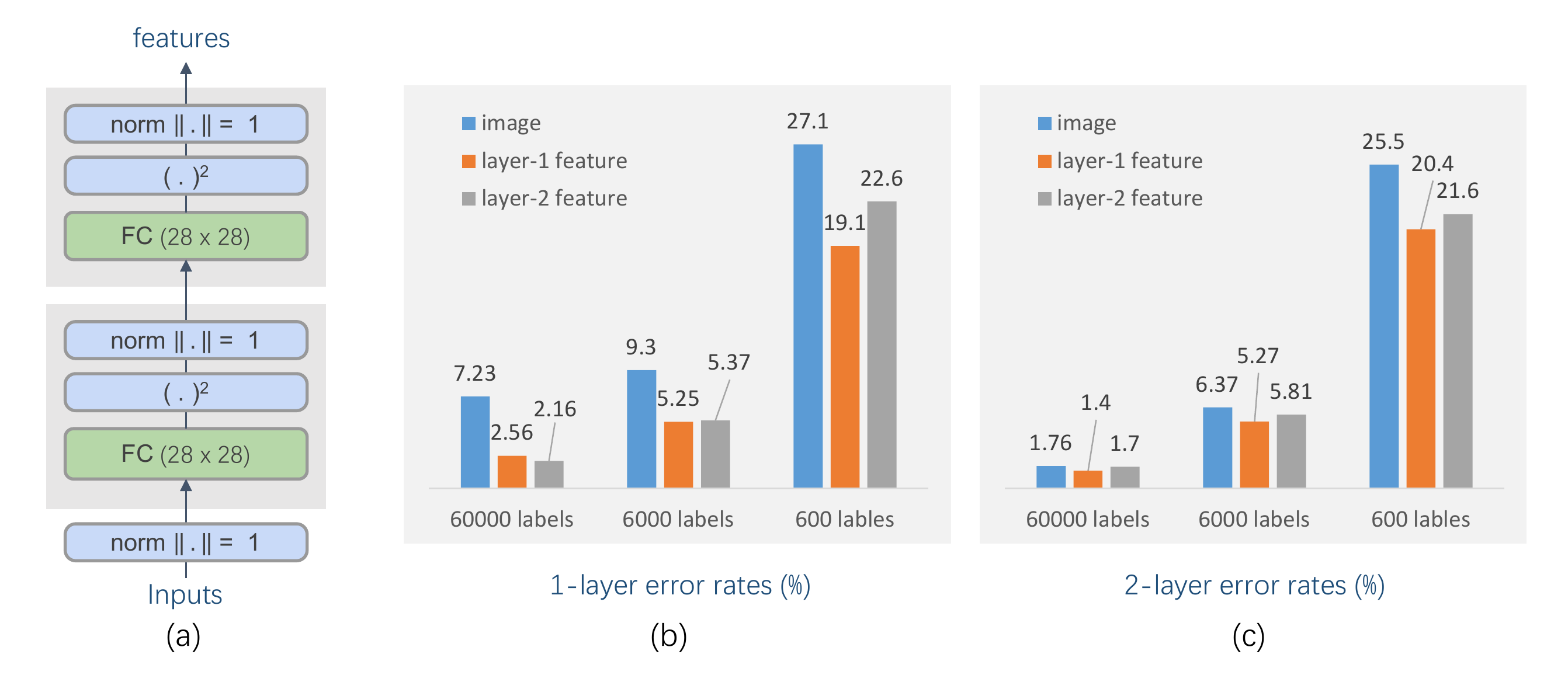}
\caption{Classification using the unsupervised features learned by the competitive learning rule on MNIST. (a) The structure of a network to learn unsupervised features. (b) The test error rates of the network with a single layer trained by backpropagation with $60000$, $6000$, or $600$ labeled images, respectively. It compares three input scenarios: normalized images, first layer unsupervised features, and second layer unsupervised features. (c) The classification error rates of a two-layer network trained by backpropagation.}
\label{fig_feature_performance}
\end{figure}

Experiments are conducted to investigate the unsupervised features learned by the local competitive learning rule (\ref{equ2}). In the experiments, a neural network of $l$ fully connected feature layers, with each layer consisting of $28\times 28$ neurons, is trained using $55000$ normalized training images without labels.
As illustrated in Fig. \ref{fig_feature_performance}(a), the square  $(\cdot)^2$ is utilized as the activation function to make strong activation stronger, and it is then normalized so that the feature of each layer is a unit vector. The network is trained layer by layer from the bottom up; after one layer's training is complete, its connection weights are locked, and the training on the next layer commences.
Fig. \ref{fig_network_feature} depicts the learned unsupervised features and compares them with those learned by backpropagation using the same neural network connected to $10$ output units at the top, whose softmax is used as the predicted probability distribution of the input digit.

To evaluate the features learned without supervision, they are provided as input to the classification task as opposed to the original images. A fully connected neural network with zero or one hidden layer and one output layer of $10$ units is used for classification. Each hidden layer consists of $28\times 28$ neurons, and the activation function is $\mathrm{ReLU}$. The softmax of the ten output units is used as the predicted probability distribution of the digit of the input image, and the classified digit is the one with the highest softmax value. The neural network used for classification is trained by backpropagation on $60000$, $6000$, and $600$ images, respectively, in which $5000$, $1000$, and $100$ images are used for validation.

Fig. \ref{fig_feature_performance}(b) shows the test error rates of a neural network without hidden layers that uses the learned features or the original images as inputs. When there are $60000$ labeled images, using the original images as inputs leads to a test error rate of $7.23\%$. Using the 1st-layer features learned by the competitive learning rule, this error rate can be cut to $2.56\%$. Given that the proposed learning rule tries to extract correlations from the input patterns independent of any task, the extracted features are generic and can be used for a wide variety of learning tasks. If we substitute the $1$st-layer unsupervised features with the $2$nd-layer unsupervised features, the accuracy of classification may decline slightly, due in part to the loss of information during feature extraction. As depicted in \ref{fig_feature_performance}(c), a similar phenomenon can be observed when using a neural network of two layers. Using $60000$ labeled images, we see that the performance of the $2$nd-layer features for classification is inferior to that of the $1$st-layer features.

\subsection{Pre-training Model}

\begin{figure}[!t]
\centering
\includegraphics[width=5.8in]{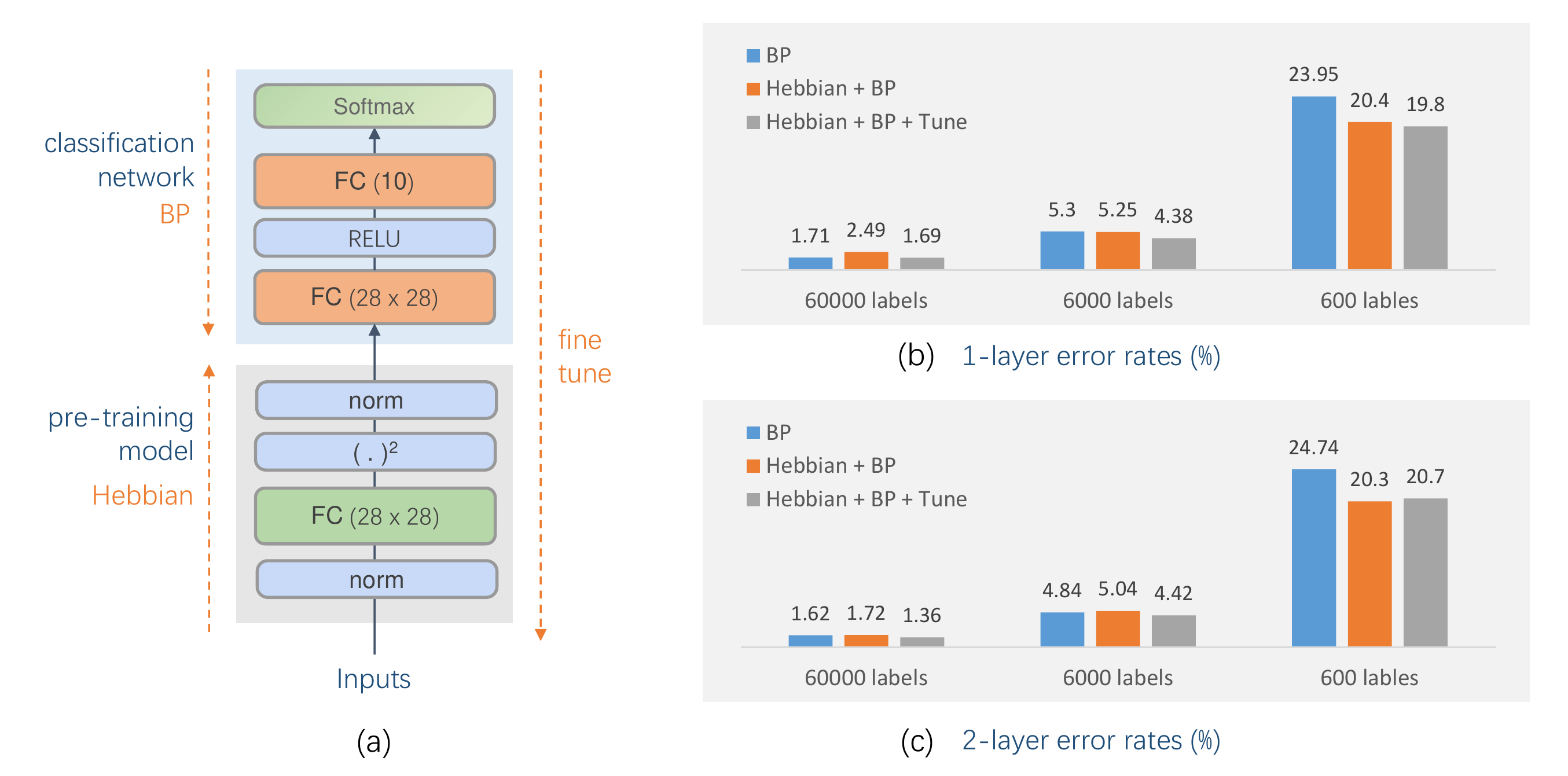}
\caption{A pre-training model based on the local competitive learning. (a) A network structure composed of a pre-training model and a classification network. (b) The test error rates of the network with  a single layer atop the pre-training model. It compares three different training methods for a predetermined number of labeled images. (c) The test error rates of the network with two layers above the pre-training model. }
\label{fig_mnist_feature}
\end{figure}

The network trained by the local competitive learning rule can also be utilized as a pre-training model that is fine-tuned in the future via backpropagation for specific learning tasks. The fine-tuning assists in reshaping the unsupervised features in order to better adapt to the target task. In the experiments, the used neural network includes two components: a pre-training model and a classification network, as shown in Fig. \ref{fig_mnist_feature}(a). Based on which, we conduct a group of experiments on MNIST and compare the performances of three different training methods: (1) training the entire network by backpropagation; (2) training the pre-training model based on all the $60000$ images without using their labels by the local competitive learning rule and then training the classification network by backpropagation; and (3) fine-tuning the entire network by propagation based on the one trained by the second method.

Figs. \ref{fig_mnist_feature}(b) and (c) represent the test error rates when the classification network contains a single layer or two layers, respectively. We see that, given all the $60000$ images labeled,  the second method leads to a higher error rate than the first method, which trains the full network by backpropagation. But when the number of labeled images is reduced to $600$, the second method leads to considerably superior performance. For almost all the experiments, the third method with fine-tuning worked the best. This indicates that the unsupervised features gained from a large amount of unlabeled data are beneficial to a supervised learning task, and supervised information can be utilized to adjust the features and enhance learning. Our goal here is not to surpass the state-of-the-art on MNIST but to show that the performance of the backpropagation can be further improved with the help of unsupervised local competitive learning.



\bibliographystyle{IEEEtran}
\bibliography{sn-bibliography}% common bib file


% \end{thebibliography}


% \end{IEEEbiography}


%\newpage



%\vfill


\end{document}